\documentclass[jou,apacite]{apa6}

\usepackage{graphicx}

\usepackage{amsmath}
\usepackage{amsfonts}
\usepackage{color}
\usepackage{array}
\usepackage{multirow}
\usepackage{subfig}

\newcommand{\RR}{\mathbb{R}}
\newcommand{\PP}{\mathbb{P}}
\newcommand{\QQ}{\mathbb{Q}}
\newcommand{\HH}{\mathbb{H}}
\renewcommand{\vec}[1]{\mathbf{#1}}

\newtheorem{theorem}{Theorem}

\title{Transport-Based Pattern Theory: A Signal Transformation Approach}
\shorttitle{Transport-Based Pattern Theory}

\twoauthors{Liam Cattell}{Gustavo K. Rohde}
\twoaffiliations{Department of Biomedical Engineering, University of Virginia, Charlottesville, USA}{Department of Biomedical Engineering, University of Virginia, Charlottesville, USA\\Charles L. Brown Department of Electrical and Computer Engineering, University of Virginia, Charlottesville, USA}

\abstract{In many scientific fields imaging is used to relate a certain physical quantity to other dependent variables. Therefore, images can be considered as a map from a real-world coordinate system to the non-negative measurements being acquired. In this work we describe an approach for simultaneous modeling and inference of such data, using the mathematics of optimal transport. To achieve this, we describe a numerical implementation of the linear optimal transport transform, based on the solution of the Monge-Amp\`{e}re equation, which uses Brenier's theorem to characterize the solution of the Monge functional as the derivative of a convex potential function. We use our implementation of the transform to compute a curl-free mapping between two images, and show that it is able to match images with lower error that existing methods. Moreover, we provide theoretical justification for properties of the linear optimal transport framework observed in the literature, including a theorem for the linear separation of data classes. Finally, we use our optimal transport method to empirically demonstrate that the linear separability theorem holds, by rendering non-linearly separable data as linearly separable following transform to transport space.}

\rightheader{Transport-Based Pattern Theory}
\leftheader{Liam Cattell}

\begin{document}
\maketitle

\section{Introduction}
\label{sec:introduction}

Many imaging problems relate to analyzing spatial patterns of a certain physical quantity for the purpose of understanding the spatiotemporal distribution of such quantity in relation to other dependent variables. Historically, astronomers, for example, used images from telescopes to understand the mass distribution of different galaxies \cite{schwarzschild1954,bertola1966}. Biologists and pathologists make use of digital images of cells in order to understand their structure as a function of different experimental conditions or disease \cite{yeung2005,wang2010}. Scientists and radiologists utilize magnetic resonance images (MRI), computed tomography (CT), and other imaging modalities, in order to understand the structure of different organs as they relate to dynamics, physiological, and other functional characteristics \cite{gorbunova2012,schmitter2015}. 

In the majority of these cases, imaging refers to the measurement of a certain physical quantity which is positive in nature. In X-ray CT, detectors measure the number of photons arriving at a certain region of the detector in a certain window of time. In magnitude reconstructed MRI, the quantity being reconstructed is the magnitude of the magnetization density. In light microscopy, detectors also measure the number of photons arriving within a certain time window. In transmission microscopy, light is attenuated and a darker pixel corresponds to an increased density in that location. In fluorescence microscopy, the amount of signal being received is proportional to the concentration of a certain fluorophore at that location. In these, and other examples, an image $I$ can be considered as a map from the domain $\Omega \in \RR^d$ representing the coordinate system in which measurements are being made, to the non-negative real numbers: $I:\Omega \rightarrow \RR^+$.

Here we describe an approach for modeling and inference on such data by considering each image in a database $\{I_1, I_2, \cdots, I_N \}$ as an instance of a given template $I_0$ `morphed' by a given function $f_k$. That is $I_k(x) \sim D_f(x)I_0(f(x))$, where $D_f$ is the determinant of the Jacobian of $f$. The association between $I_k$ and $f_k$ is made unique by the mathematics of optimal transport, as explained in \cite{wang2013,kolouri2016b}. The embedding provided by the so called linear optimal transport (LOT) approach can be viewed as mathematical image transforms with forward and inverse operations, thus enabling inference and modeling simultaneously. Here we extend the work presented in \cite{wang2013,kolouri2016b} to provide theoretical justification for certain properties observed empirically in the analysis of cell and brain image data bases \cite{ozolek2014,kolouri2016b,tosun2015,kundu2017}, including a linear separation theorem in transform space similar to the work shown in \cite{park2017,kolouri2016c}.

We note that the framework discussed above, whereby images are modeled as actions $D_f s_0 \circ f$ has similarities to the \emph{pattern theory} approach of Grenander and Mumford, and others \cite{grenander1994,grenander1996,grenander1998,mumford1997}. In particular, the pattern theory-based computational anatomy approach originally proposed by Grenander and Miller \cite{grenander1998} involves viewing the orbit formed by the action of a diffeomorphism as a generative model for imagery. As such, it attempts to model each observed image as $I_k(x) \sim I_0(f(x))$ based on representations, probabilities, and inference methods that can be developed within a Bayesian perspective. The framework has been useful for motivating a number of methods used in pattern recognition \cite{grenander2007}. The LOT approach introduced in \cite{wang2013,kolouri2016b,park2017} shares a similar point of view, and can be seen as a particular version of the more general pattern theory approach \cite{grenander1994,grenander1996,grenander1998,mumford1997}, with the main innovation being that the transport formulation can encompass both displacement (geometric transformations) as well as photometric (i.e. intensity differences) information.

\subsection{Overview}
\label{subsec:contributions}
We begin with Section \ref{sec:optimal-transport}, Definitions and Preliminaries, by introducing the basics of the mathematics of optimal mass transport, including the Monge functional, and Brenier's theorem that characterizes the solution of the Monge functional. We also review basic notions related to the geometry of optimal transport, as well as the LOT metric described in \cite{wang2013,kolouri2016b}. In Section \ref{sec:transport-transforms}, we define the basics of the LOT transform, including the forward and inverse operations. In Section \ref{sec:lot-properties} we describe several mathematical properties of the LOT transform. These include translation, scaling, and composition claims, as well as linear separability properties of the transform.  Section \ref{sec:spot} describes the numerical discretization and implementation of the LOT transform based on the solution of the Monge-Amp\`{e}re equation, within a multi-resolution framework. Section \ref{sec:experiments} describes computational experiments and concluding remarks are shown in Section \ref{sec:discussion}.

\section{Definitions and preliminaries}
\label{sec:optimal-transport}

\subsection{Optimal Mass Transport}
\label{subsec:ot-formulation}


Consider two Borel probability measures $\mu$ and $\nu$ with finite `p' moments defined on measure spaces $X \subseteq \RR^d$ and $Y \subseteq \RR^d$, respectively. Consider the case when the measures are smooth and are associated with positive density functions $I_0$ and $I_1$, such that $d\mu = I_0(x)dx$ and $d\nu(y) = I_1(y)dy$. The original optimal transport problem posed by Monge \cite{monge1781} is to find a mass-preserving (MP) map $f : X \rightarrow Y$ that pushes $\mu$ on to $\nu$ and minimizes the objective function

\begin{equation}	
    M(\mu,\nu) = \inf_{ f\in \textrm{MP}} \int_X c(x,f(x)) I_0(x) d(x)
    \label{eq:monge-objective}
\end{equation}
where $c:X\times Y\rightarrow \RR^+$ is the cost function, and $\textrm{MP}:= \{ f:X \rightarrow Y ~|~ f_{\#}\mu=\nu \}$, where $f_{\#}\mu$ denotes the push-forward of measure $\mu$. This is characterized as:

\begin{equation}
    \int_{f^{-1}(A)} d\mu(x) =\int_{A} d\nu(y) \quad \forall A \subset Y
    \label{eq:mp-map}
\end{equation}
If $f$ is a smooth, one-to-one mapping, Eq. \eqref{eq:mp-map} can be written in a differential form:

\begin{equation}
    D_f(x)I_1(f(x)) = I_0(x)
    \label{eq:jacobian-equation}
\end{equation}
where $D_f$ is the determinant of the Jacobian of $f$.

In his original paper, Monge considered the Euclidean distance $c(x,f(x)) = \|f(x)-x\|$ as the cost function \cite{monge1781}. However, under the smoothness assumption required for Eq. \eqref{eq:jacobian-equation}, the cost function is often chosen to be the $L^2$-norm, such that Eq. (\ref{eq:monge-objective}) becomes the $L^2$ Wasserstein metric \cite{haker2004,trigila2016}:

\begin{multline}
    M(\mu,\nu) = \inf_{ f\in \textrm{MP}} \int_X \|x-f(x)\|^2 I_0(x) dx\\
    \textrm{s.t.} ~ D_f(x)I_1(f(x)) = I_0(x)
    \label{eq:monge-problem}
\end{multline}

It should be noted that both the objective function and the constraint in Eq. \eqref{eq:monge-problem} are nonlinear with respect to $f(x)$. Moreover, an important result in the field is Brenier's theorem, which characterizes the unique solution to the Monge problem \cite{brenier1991}.

\begin{theorem}[Brenier's Theorem]
    \label{theorem:brenier}
    Let $I_0$ and $I_1$ be non-negative functions of same total mass and with bounded support:
    
    \begin{equation*}
        \int_X I_0(x)dx = \int_Y I_1(y)dy.
        \label{eq:equal-masses}
    \end{equation*}
    When $c(x,y) = h(x-y)$ for some strictly convex function $h$, then there exists a unique optimal transport map $f^*$ that minimizes Eq. \eqref{eq:monge-objective}. Furthermore, if $c(x,y) = \|x-y\|^2$ then there exists a (unique up to adding a constant) convex function $\phi$ such that $f^*(x) = \nabla \phi(x)$. A proof is available in \cite{villani2008,gangbo1996}. 
\end{theorem}

Finally, it should be mentioned that, for certain measures, the Monge formulation of the optimal transport problem is ill-posed; in the sense that there is no transport map to rearrange one measure to another. In such cases the Kantorovich formulation of the problem using transport plans is preferred. For brevity, we omit a detailed discussion of the Kantorovich mass transport formulation and instead refer the reader to \cite{kolouri2017} for more details.

\subsection{Geometry of Optimal Transport}
\label{subsec:ot-geometry}

Brenier's theorem states that the OT map between $\mu$ and $\nu$ is unique and, in the continuous setting presented above, the mass from each point $x$ is moved to a single location, given as the value of the function $f(x)$. Formally, the set of measures not just a metric space, but is also a Riemannian manifold equipped with an inner product on the tangent space at any point \cite{docarmo1992}. Specifically, the tangent space at the measure $\sigma$ with density $\gamma$ (i.e. $d\sigma = \gamma(x) dx$) is the set of the following vector fields $T_\sigma = \{ v: \Omega \to \RR^d \text{ such that } \int_\Omega |v(x)|^2 \gamma(x) dx < \infty \}$ and the inner product is the weighted $L^2$ \cite{wang2013}:

\begin{equation*}
    \langle v_1, v_2 \rangle_\sigma = \int_\Omega v_1(x) \cdot v_2(x) I_0(x) dx
\end{equation*}

Therefore, the OT distance is the length of the geodesic connecting two measures \cite{benamou2000}. Fortunately, in the context of optimal transport, this has a straightforward interpretation: if $\mu_t$, $0 \leq t \leq 1$ is the geodesic connecting $\mu$ to $\nu$, then $\mu_t$ is the measure obtained when mass from $\mu$ is transported by the transportation map $x \to (1-t)x + t f(x)$. Consequently, $\mu_t(A) = \int_{f_t^{-1}(A)} \alpha(x) dx$.

\subsection{The Linear Optimal Transport Distance}
\label{subsec:ot-embeddings}

Computational complexity can hinder the application of OT-related metrics to learning and pattern analysis problems that involve large numbers of images. For recognition problems, the application of classification techniques such as nearest neighbors \cite{altman1992}, linear discriminant analysis \cite{mclachlan2004,wang2011}, support vector machines \cite{cortes1995} and their respective kernel versions \cite{cristianini2000}, would require $O(N^2)$ OT computations, with $N$ representing the number of images in the dataset. To ease this burden, Wang et al. \cite{wang2013} proposed the linear optimal transport (LOT) metric, which borrows the statistical atlas notion often seen in brain morphology studies \cite{ashburner2000}, and seeks to compare images to one another based on their relation to a reference.

Consider a given signal $I$ associated with measure $\mu$ via $d\mu(y) = I(y)dy$, $y \in Y$ as before, and a reference function $I_0$ associated with measure $\sigma$ such that $d\sigma(x) = I_0(x)dx$, $x \in X$.  We can then identify $\mu$ (and $I$) with a map $f$ via the optimization problem stated in Eq. \eqref{eq:monge-objective}. Brenier's theorem tells us that this identification is unique and that it is characterized by the map being the gradient of a potential function $\phi$, that is $f = \nabla \phi$, such that $D_f(x)I(f(x)) = I_0(x)$. This identification can be viewed as the projection of $\mu$ on to the tangent plane $T_\sigma$ described above. More precisely, let the projection of $\mu$ be:

\begin{equation*}
    P(\mu) = v \: \textrm{ where } \:  v(x) = f(x) - x.
\end{equation*}

Then $v \in T_\sigma$:

\begin{equation}
     \int_X |v(x)|^2 I_0(x) dx  = \langle v , v \rangle_\sigma = \|P(\mu) - P(\sigma)\|_\sigma^2, \nonumber
\end{equation}
where $\|v \|_\sigma^2$ is defined to be $\langle v,v \rangle_\sigma$. As explained in \cite{wang2013}, this is known as an azimuthal equidistant projection in cartography, while in differential geometry it would be the inverse of the exponential map. Finally, given two measures $\mu$ and $\nu$, and a fixed reference $\sigma$, the LOT distance between $\mu$ and $\nu$ is defined as \cite{wang2013}:

\begin{equation}
    \label{eq:lot_cont}
    d_{LOT}(\mu, \nu) = \| P(\mu) - P(\nu) \|_\sigma. 
\end{equation}

\section{LOT Transform}
\label{sec:transport-transforms}

Consider two measures $\mu_1$ and $\mu_2$, with corresponding densities  $I_1$ and $I_2$, and let $\sigma$ be a fixed reference measure with corresponding density $I_0$. Furthermore, let $f_1$ and $f_2$ correspond to the OT maps linking $\mu_1$ to $\sigma$ and $\mu_2$ to $\sigma$. That is $D_{f_1}(x)I_1(f_1(x)) = D_{f_2}(x)I_2(f_2(x)) = I_0(x)$ such that $f$ is uniquely defined through the solution to the Monge minimization problem in Eq. \eqref{eq:monge-objective}. From Eq. \eqref{eq:lot_cont} we have:
\begin{eqnarray*}
    d^2_{LOT}(\mu_1,\mu_2) &=& \| P(\mu) - P(\nu) \|_\sigma^2\\
    &=& \int_{X} |(f_1(\vec{x})-x) - (f_2(\vec{x})-x)|^2 I_0(\vec{x}) d\vec{x}\\
    &=& \| \hat{I}_1 - \hat{I}_2 \|^2
\end{eqnarray*}
where $\| \hat{I} \|^2$ is defined to be $\langle \hat{I},\hat{I} \rangle$, and
\begin{eqnarray*}
    \hat{I}_1 = (f_1(x)-x)\sqrt{I_0(x)},\\
    \hat{I}_2 = (f_2(x)-x)\sqrt{I_0(x)}.
\end{eqnarray*}

The result above motivated Kolouri \cite{kolouri2016} to define the continuous LOT embedding (transform) for signals and images. Here, a signal is interpreted to be a continuous function $I(y)$, with $Y \in \mathbb{R}^d$, where $d$ refers to the dimension of the signal (i.e. $d=1$ for 1D signals, $d=2$ for 2D images, etc.). The forward (analysis) LOT transform of $I$ can then be defined as:

\begin{equation}
    \hat{I}(x) = (f(x) - x) \sqrt{I_0(x)}, \quad x \in X
    \label{eq:lot_forward}
\end{equation}
where $f$ is the gradient of a potential function $\phi$, such that $f = \nabla \phi$, and $D_{f}(x)I(f(x)) = I_0(x)$. The inverse (synthesis) LOT transform of $\hat{I}$ is then:
\begin{equation}
 I(y) = f^{-1}(y) I_0(f^{-1}(y)), \quad y \in Y
 \label{eq:lot_inverse}
\end{equation}
where $f^{-1}(f(x)) = x$. Thus, we can consider the LOT transform as an operation taking signals from one domain (signal space) to another (displacements). Just like in engineering, where we consider time/space domain versus frequency domain, the LOT transform allows us to consider signal space domain versus displacement domain (with respect to a fixed reference).

The idea was extended by Park et al. \cite{park2017}, who exploited the fact that, in 1D, the unique map between any two positive smooth densities can be computed in closed form. Consequently, the authors defined the Cumulative Distribution Transform (CDT) for 1D signals \cite{park2017}. This allowed for the definition of several interesting transform pair properties, some of which have relevance to learning problems that are popular at the time of writing. Kolouri et al. \cite{kolouri2016b} expanded the CDT to higher dimensional signals (images) by using the Radon transform formalism to obtain a set of 1D projections from an image, and applying the CDT along these projections. This is equivalent to obtaining an embedding for the sliced Wasserstein metric \cite{kolouri2016c}. Similarly to the LOT transform above, the CDT and Radon-CDT are invertible operations with forward (analysis) and inverse (synthesis) operations. 





\section{LOT Transform Properties}
\label{sec:lot-properties}

In this section, we present three properties of the LOT transform pertaining to changes in density coordinates: the translation theorem, scaling theorem, and composition theorem. In addition, we demonstrate that the linear separation property of the CDT and Radon-CDT also holds for the LOT transform in $d$-dimensions.

\subsection{Translation}
\label{subsec:lot-translation}

Let $I_\mu$ be the translation of the density $I_1$ by $\mu$, such that $I_\mu(y) = I_1(y-\mu)$. The transform of $I_\mu$ with respect to the reference density $I_0$ is given by:

\begin{equation}
    \hat{I}_\mu(x) = \hat{I}_1(x) + \mu \sqrt{I_0(x)} \quad x \in X
    \label{eq:lot-translation}
\end{equation}
where $I_0$ is defined on measure space $X \subseteq \RR^d$. For a proof, see Appendix \ref{app:proof-translation}.

\subsection{Scaling}
\label{subsec:lot-scaling}

Let $I_\sigma$ be the scaling of the density $I_1$ by $\sigma$, such that $I_\sigma(x) = \sigma I_1(\sigma x)$. The transform of $I_\sigma$ with respect to the reference density $I_0$ is given by:

\begin{equation}
    \hat{I}_\sigma(x) = \frac{1}{\sigma} \left( \hat{I}_1(x) - x(\sigma - 1)\sqrt{I_0(x)} \right) \quad x \in X
    \label{eq:lot-scaling}
\end{equation}
where $I_0$ is defined on measure space $X \subseteq \RR^d$. For a proof, see Appendix \ref{app:proof-scaling}.

\subsection{Composition}
\label{subsec:lot-composition}

Let $I_g$ represent the composition of the density $I_1$ with an invertible function $g$, such that $I_g(x) = D_g(x) I_1(g(x))$, where $D_g$ is the determinant of the Jacobian of $g$, and $g$ is the gradient of a convex potential function $\varphi$ (i.e. $g = \nabla \varphi$). The transform of $I_g$ with respect to the reference density $I_0$ is given by:

\begin{equation}
    \hat{I}_g(x) = \left( g^{-1} \left( \frac{\hat{I}_1(x)}{\sqrt{I_0(x)}} + x \right) - x \right) \sqrt{I_0(x)} \quad x \in X
    \label{eq:lot-composition}
\end{equation}
where $I_0$ is defined on measure space $X \subseteq \RR^d$. For a proof, see Appendix \ref{app:proof-composition}.

\subsection{Linear Separability in LOT Space}
\label{subsec:lot-linear-separability}

It is well-known that if two sets are convex and disjoint, there always exists a hyperplane that is able to separate the sets (see \cite{park2017}, for example). As a consequence, the sets are said to be linearly separable. In this section we will demonstrate that, under certain conditions, the LOT transform can enhance the linear separability of image classes.


Consider two non-overlapping, non-linearly separable subsets $\PP$ and $\QQ$ of a vector space $V$. All elements $p_i$ of $\PP$ are generated by transforming a ``mother'' density $p_0$ with a differentiable function $h_i$:

\begin{equation*}
    p_i = D_{h_i} p_0 \circ h_i \quad p_i \in \PP, \: h_i \in \HH
\end{equation*}
where $h$ is the gradient of a convex potential $\phi$, such that $h = \nabla \phi$. Moreover, $D_h$ denotes the determinant of the Jacobian of $h$, $p_0 \circ h_i(x) = p_0(h_i(x))$ and $\HH$ is a set of diffeomorphisms (e.g. the set of all translations). Similarly, the elements $q_j$ of $\QQ$ are generated by transforming $q_0$ with a differentiable function $h_j \in \HH$:

\begin{equation*}
    q_j = D_{h_j} q_0 \circ h_j \quad q_j \in \QQ, \: h_j \in \HH
\end{equation*}
It should also be noted that since sets $\PP$ and $\QQ$ do not overlap:

\begin{equation*}
    D_{h} p_0 \circ h \neq q_0 \quad \forall h \in \HH
\end{equation*}

The definitions above provide a framework that can be used to construct or interpret image classes. For example, let $\HH$ be the set of all translations $h(x) = x - \mu$, where $\mu$ is a random variable. The elements of sets $\PP$ and $\QQ$ are given by $p_0 \circ h$ and $q_0 \circ h$, respectively, and are translations of the original ``mother'' densities $p_0$ and $q_0$. Therefore, the translation $\mu$ becomes the confound parameter that a classifier must decode in order to separate the classes.

\vspace{0.1in}
\begin{theorem} {\em (Linear separability in LOT space)}
Let $\PP,\, \QQ,\, \HH$ follow the definitions given above. If $\hat{\PP}$ and $\hat{\QQ}$ are the LOT transforms of $\PP$ and $\QQ$, then $\hat{\PP}$ and $\hat{\QQ}$ will be linearly separable if $\HH$ satisfies the following conditions:

\renewcommand{\labelenumi}{\roman{enumi})}
\begin{enumerate}
    \item $\forall h \in \HH, \quad h^{-1} \in \HH$
    \item $\forall h \in \HH, \quad \sum_i \alpha_i h_i \in \HH$ where $\sum_i \alpha_i = 1$
    \item $\forall h_1, h_2 \in \HH, \quad h_1 \circ h_2 \in \HH$
\end{enumerate}
\end{theorem}

For a proof, see Appendix \ref{app:proof-linear-separability}.
\vspace{0.1in}

We note that the definition of the embedding (transform) for a certain image $I$  given in Eq. \eqref{eq:lot_forward} is in a way redundant and that the convex potential $\phi$ could be used instead. In fact, doing so would be preferred in many practical situations, such as performing machine learning-type operations on digital images. This is because the array used to store the transform defined in Eq. \eqref{eq:lot_forward} would be twice the size of the array used to store the potential $\phi$, since there would be derivatives in the $x-$ and $y-$directions. More than just a savings in computer memory, using the potential directly would translate to performing operations on objects of lower dimension. Ultimately, the mathematical meaning is the same, as the two formulations are related by a gradient operation. However, gradient operations are also known to be noise augmenting, and thus introduce additional difficulties when dealing with real data. Thus in the computational examples we utilize the potential $\phi$, rather than the LOT defined in Eq. \eqref{eq:lot_forward}.






\section{Numerical Implementation}
\label{sec:spot}

Given image $I:\Omega\rightarrow \mathbb{R}^+$, and reference $I_0:\Omega\rightarrow \mathbb{R}^+$, with $\Omega = [0,1]^2$ (in the results shown below we use 2D images), we compute an approximate solution for the the LOT transform by solving the associated Monge-Amp\`{e}re partial differential equation. Since Brenier's theorem asserts that the optimal map between $I$ and $I_0$ is characterized by $f = \nabla \phi$, we can write:
\begin{equation}
    \det H \phi(x) = \frac{I_0(x)}{I(\nabla \phi(x))}
\end{equation}
where $\det H \phi(x)$ represents the determinant of the Hessian matrix of second partial derivatives of $\phi$. In our implementation we use a parametric formulation for the potential function based on a linear combination of basis functions $\rho$:

\begin{equation}
    \phi(x) = \frac{1}{2}|x|^2+ \sum_{p \in Z} c(p) \rho(x-p)
    \label{eq:potential}
\end{equation}
where $c$ are the coefficients (to be determined via optimization), and $Z$ is the pixel grid for the image. For simplicity, we chose the basis function $\rho$ to be a radially symmetric Gaussian of standard deviation $\sigma$, where $\sigma$ is a user-defined parameter:

\begin{equation}
    \rho(x) = \frac{1}{\sqrt{2\pi \sigma^2}} e^{-x^2/(2\sigma^2)}
    \label{eq:gaussian}
\end{equation}
By combining Brenier's theorem with Eq. \eqref{eq:potential}), the model for the spatial transformation is then given by:

\begin{equation}
    f(x) = x - \sum_{p \in Z} c(p) \nabla \rho(x-p).
    \label{eq:spot-spatial-transformation}
\end{equation}
Rather than solving the Monge-Amp\`{e}re PDE directly, we propose to approximate the solution by iterative optimization of the related cost function (presented in discretized form):

\begin{equation}
    \psi(c) = \frac{1}{2} \sum_{p \in Z} \big( D(p)I_1(f(p)) - I_0(p) \big)^2 dx
    \label{eq:spot-objective}
\end{equation}
Thus, the solution for the OT problem can be found by solving a registration-like image processing problem, under the model for the spatial transformation specified in Eqs. \eqref{eq:gaussian} and \eqref{eq:spot-spatial-transformation}. Note that for the spatial transformation to remain invertible, $D_f(x) = H\phi$ must be positive definite throughout the domain of the image, and thus input images must be such that $I_0 > 0, I_1 > 0$. Finally, we note the approach has similarities to other parametric image registration methods based on basis functions \cite{rohde2003,kybic2003,rueckert1999}.


\subsection{Numerical Optimization}
\label{subsec:optimization}

The objective function $\psi$ can be minimized using gradient descent, which can be described by the following iterative update rule:

\begin{equation}
    c_{t+1} = c_t - \eta \psi_c(c_t)
\end{equation}
where $\eta$ is the learning rate and $\psi_c$ is the derivative of $\psi$ with respect to $c$. We refer the reader to Appendix \ref{app:spot-objective} for the full derivation of $\psi_c$ for the case when $I_0$ and $I_1$ are 2D images.

It is well known that standard gradient descent can be very slow to converge to the optimum, and therefore, alternative methods have been developed that can dramatically increase the rate of convergence \cite{nesterov1983,duchi2011,kingma2015}. In this work, we use the method known as Adam \cite{kingma2015}, which typically requires little tuning and has been shown to outperform other first-order gradient-based methods in neural networks. The Adam method estimates the first moment $m_t$ and second moment $v_t$ of the gradient:

\begin{equation}
    \begin{aligned}
        m_t &= \beta_1 m_{t-1} + (1-\beta_1)\psi_c(c_t)\\
        v_t &= \beta_2 v_{t-1} + (1-\beta_2)\psi_c^2(c_t)
    \end{aligned}
\end{equation}
where $\beta_1, \beta_2 \in [0,1)$ are user-defined exponential decay rates.

Since $m_t$ and $v_t$ are initialized as zero, the moment estimates are biased towards zero. Kingma and Ba \cite{kingma2015} counteract this problem by computing bias-corrected estimates $\hat{m}_t$ and $\hat{v}_t$:

\begin{equation}
    \begin{aligned}
        \hat{m}_t = \frac{m_t}{1-\beta_1^t}\\
        \hat{v}_t = \frac{v_t}{1-\beta_2^t}\\
    \end{aligned}
\end{equation}

These estimates are then used to compute an adaptive learning rate. As a result, the coefficients $c$ are updated as follows:
\begin{equation}
    c_{t+1} = c_t - \frac{\eta}{\sqrt{\hat{v}_t} + \epsilon} \hat{m}_t
\end{equation}

The authors suggest default values of $\beta_1 = 0.9$, $\beta_2 = 0.999$, and $\epsilon = 10^{-8}$ \cite{kingma2015}. We adopt these values throughout this work.

\subsection{Multi-Scale Approach}
\label{subsec:multi-scale}

The efficiency of the minimization not only relies on the choice of gradient descent algorithm, but also the position from where the optimization is started. Generally, we can assume that an image comprises a hierarchy of scales. Low frequency features, such as the general shapes within the image, are found at the coarsest scale, and high frequency features, such as noise, are found at the finest scale \cite{paquin2006}. We can guide our method towards the globally optimum potential $\phi^*$, by computing the optimum potential $\phi_i^*$ at scale $i$, and using the result as the initial position for the gradient descent at the next, finer scale. For $N$ scales, the globally optimum potential $\phi^*$ can be expressed as follows:

\begin{equation}
    \phi^*(x) = \sum_{i=1}^N \phi_i^*(x)
\end{equation}

By extension, the optimal transport map $f^*$ is given by:

\begin{equation}
    f^*(x) = x - \sum_{i=1}^N \nabla\phi_i^*(x)\\
\end{equation}

\begin{figure}[t!]
    \centering
    \includegraphics[width=84mm]{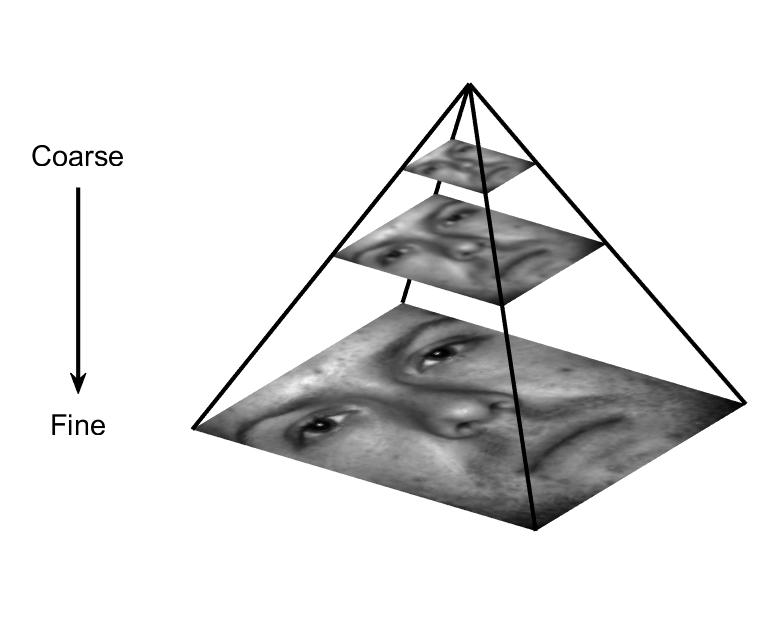}
    \caption{An illustration of a Gaussian pyramid used in the multi-scale version of our single potential optimal transport method.}
    \label{fig:gaussian-pyramid}
\end{figure}

In this work, we generated a hierarchy of image scales using a Gaussian pyramid. The Gaussian pyramid is constructed by smoothing the images $I_0$ and $I_1$ with a Gaussian filter, and subsampling the smoothed images by a factor of two in each direction \cite{adelson1984}. An illustration of the result is depicted in Fig. \ref{fig:gaussian-pyramid}. During the multi-scale gradient descent, moving from a coarse scale to a finer scale requires upsampling the potential $\phi_i^*$. Unlike the multi-scale approach developed by Kundu et al. \cite{kundu2017}, which requires the transport maps to be interpolated from one scale to the next, the potential $\phi_i^*$ in our method can be upsampled using Eq. (\ref{eq:potential}) with appropriate values for $x$.

\section{Experiments}
\label{sec:experiments}

\subsection{Image Warping}
\label{subsec:exp-registration}

\subsubsection{Data Preparation}
\label{subsubsec:reg-data-preparation}

We tested our proposed single potential optimal transport method on three separate datasets: the extended Yale Face Database B \cite{lee2005}, the LONI Probabilistic Brain Atlas dataset \cite{shattuck2008}, and a random sample of images from ImageNet \cite{deng2009}. In this section, we briefly describe the datasets and any pre-processing that was conducted prior to our experiments.

\paragraph{YaleB}
\label{par:yaleb}

\begin{figure}[t!]
    \centering
        \subfloat[YaleB]{\includegraphics[width=84mm]{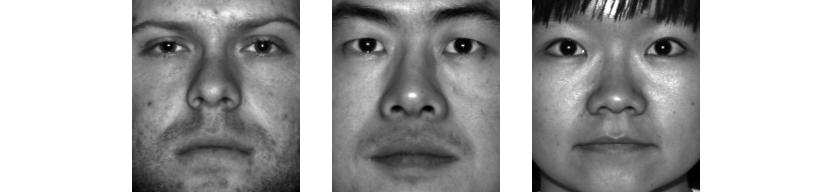}%
        \label{fig:data-yaleb}}
    \hfill
        \subfloat[LPBA40]{\includegraphics[width=84mm]{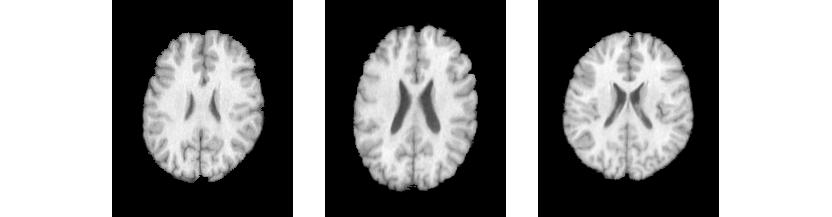}%
        \label{fig:data-lpba40}}
    \hfill
        \subfloat[ImageNet]{\includegraphics[width=84mm]{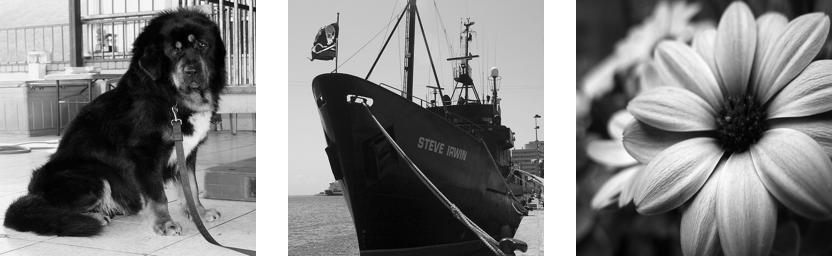}%
        \label{fig:data-imagenet}}
    \caption{Sample images after pre-processing from \protect\subref{fig:data-yaleb} the YaleB dataset, \protect\subref{fig:data-lpba40} the LPBA40 dataset, and \protect\subref{fig:data-imagenet} the ImageNet dataset.}
    \label{fig:reg-data}
\end{figure}

The extended Yale Face Database B \cite{lee2005} (abbreviated to YaleB) comprises 1710 grayscale images of 38 individuals under 45 different lighting conditions. The authors of the dataset aligned the images to one another using the location of the eyes as landmarks. The images were also cropped to exclude hair and ears \cite{lee2005}. In order to limit the number of image matches, we used a single cropped image from each subject under direct illumination, resulting in a dataset of 38 images. Exemplar images from our reduced YaleB dataset are shown in Fig. \ref{fig:data-yaleb}. Prior to performing the warpings, the images were normalized so that the sum of intensities was the same in every image. This ensured that the total mass constraint in Brenier's theorem was satisfied. Identically to Kundu et al. \cite{kundu2017}, we normalized the images to a large value ($10^6$) in order to avoid numerical precision errors due to small numbers. We also added a small constant (0.1) to the normalized images so that they were strictly positive.

\paragraph{LPBA40}
\label{par:lpba40}

Brain image data from 40 subjects were used by the Laboratory of Neuro Imaging (LONI) at UCLA to construct the LONI Probabilistic Brain Atlas (LPBA) \cite{shattuck2008}. The three-dimensional magnetic resonance images were skull-stripped and aligned to a standard space using rigid-body transformations \cite{klein2009}. In this work, we used the central axial slice from the 40 pre-processed volumes, and normalized these 2D slices in an identical manner to the YaleB data. Exemplar LPBA40 images are shown in Fig. \ref{fig:data-lpba40}.

\paragraph{ImageNet}
\label{par:imagenet}

The ImageNet database consists of 3.2 million color images spread over 5247 different categories \cite{deng2009}. The large dataset is typically used to train and test machine learning methods for object detection and recognition. However, in our work, we used a randomly selected subset of 40 images from the database. Prior to use in our experiments, we converted the subset of ImageNet images to grayscale and resized them to $256\times256$ pixels. The images were normalized using the same procedure as the YaleB and LPBA40 data. Sample ImageNet images after pre-preprocessing are shown in Fig. \ref{fig:data-imagenet}.

\subsubsection{Warping Experiments}
\label{subsubsec:reg-experiments}

As indicated in Section \ref{sec:spot}, our single potential optimal transport (SPOT) method has three user-defined parameters: the standard deviation $\sigma$ of the Gaussian basis function, the learning rate $\eta$ for the gradient descent, and the number of scales $N$. In order to assess the effect of computing the optimal transport maps using a multi-scale approach, we compared SPOT at a single image scale (i.e. $N=1$) and SPOT with $N=3$ (henceforth called multi-SPOT).

To determine the optimum values for $\sigma$ and $\eta$, we split each dataset into a training and test set. The three training sets comprised five images randomly selected from the corresponding dataset. For each of the training sets, we computed the optimal transport map from every image to every other image in the set for a range of parameter values. This produced 20 ($5^2 - 5$) transport maps for each set of parameters. The parameters that resulted in the lowest mean value of the objective function (Eq. (\ref{eq:spot-objective})) across the 20 mappings for a given dataset were used to test (multi-)SPOT on that dataset.

The SPOT and multi-SPOT methods were tested using the data excluded from the parameter optimization. For each of the three test datasets, we computed the optimal transport map from every image to every other image in the set. This resulted in 1056 ($33^2 - 33$) transport maps for the YaleB dataset, and 1190 ($35^2 - 35$) transport maps for the LPBA40 and ImageNet datasets.

\subsubsection{Evaluation Measures}
\label{subsubsec:reg-evaluation-measures}

Following the computation of the transport maps, we evaluated the SPOT and multi-SPOT methods by measuring the relative mean squared error (MSE) between $D_f(x)I_1(f(x))$ and $I_0$, the proportion of mass transported, and the mean absolute curl of the optimal transport map. The measures are defined as follows:

\begin{equation}
    \textrm{Relative MSE} = \frac{\int_X \big( D(x)I_1(f(x)) - I_0(x) \big)^2 dx}{\int_X \big(I_1(x) - I_0(x) \big)^2 dx}
\end{equation}

\begin{equation}
    \textrm{Mass transported} = \frac{\int_X \|f(x)-x\|^2 I_0(x) dx}{\int_X I_0(x) dx}
\end{equation}

\begin{equation}
    \textrm{Mean absolute curl} = \frac{1}{\textrm{vol}(X)} \int_X \| \nabla \times f(x) \| dx
\end{equation}

For all three measures, the smaller the value, the closer the computed map is to the true OT map.

In order to assess whether the relative MSEs of the SPOT and multi-SPOT methods were significantly different from one another, we performed a paired $t$-test on the relative MSE results. Furthermore, we also performed paired $t$-tests for the mass transported and mean absolute curl.

\subsubsection{Comparison to Other Optimal Transport Methods}
\label{subsubsec:reg-comparison}

We compared our single-scale SPOT method to the three methods outlined in the following sections: the flow minimization (FM) method by Haker et al. \cite{haker2004}, the dual problem minimization (DPM) method by Chartrand et al. \cite{chartrand2009}, and a single-scale version of the variational optimal transport (VOT) method by Kundu et al. \cite{kundu2017}. We also compared our multi-SPOT method to the multi-scale variational optimal transport method (multi-VOT) \cite{kundu2017}. The optimum parameters for the FM, DPM, VOT, and multi-VOT methods were determined using the approach described in Section \ref{subsubsec:reg-experiments}. We then repeated the image warping experiments for all of the optimal transport methods, and evaluated the resulting transport maps using the measures defined in Section \ref{subsubsec:reg-evaluation-measures}.

\paragraph{Flow Minimization (FM)}
\label{par:haker}

Haker et al. \cite{haker2004} proposed a flow minimization method to compute the optimal transport map from the Monge formulation of the optimal transport problem. The method first obtains an initial transport map $f_0$ using the Knothe-Rosenblatt rearrangement \cite{rosenblatt1952,knothe1957}. If the initial map $f_0$ is considered to be a smooth vector field, Helmholtz's theorem can be used to resolve $f_0$ into the sum of a curl-free and a divergence-free vector field:

\begin{equation*}
    f_0 = \nabla\phi - \chi
\end{equation*}

The method by Haker et al. \cite{haker2004} uses the Helmholtz decomposition of $f_0$, and removes the curl by minimizing the objective function in Eq. (\ref{eq:monge-objective}) with cost function $c(x,f(x)) = \|f(x) - x\|^2$. The minimization is achieved using a gradient descent scheme:

\begin{equation*}
    f_{t+1}(x) = f_t(x) + \eta\frac{1}{I_0} D(x) \big(f_t(x) - \nabla (\Delta^{-1} \nabla\cdot f_t(x)) \big)
\end{equation*}
where $\Delta$ is the Laplace operator and $\nabla\cdot (\cdot)$ denotes the divergence operator.

\paragraph{Minimizing the Dual Problem (DPM)}
\label{par:chartrand}

To compute the optimal transport map between two images Chartrand et al. \cite{chartrand2009} use Kantorovich's dual form of the problem \cite{kantorovich1948}:

\begin{equation*}
    K_D(u,v) = \sup_{u,v} \int_X u(x)I_0(x)dx + \int_Y v(y)I_1(y)dy
    \label{eq:kantorovich-dual-objective}
\end{equation*}

over all integrable functions $u$ and $v$ satisfying

\begin{equation*}
    u(x) + v(y) \leq c(x,y)
\end{equation*}

By considering the cost as a strictly convex function $c(x,y) = \frac{1}{2}\|x-y\|^2$, and redefining $u(x) \rightarrow \frac{1}{2}\|x\|^2 - u(x)$ and $v(x) \rightarrow \frac{1}{2}\|y\|^2 - v(y)$, the resulting problem is to minimize:

\begin{equation*}
    K_D(u,v) = \int_X u(x)I_0(x)dx + \int_Y v(y)I_1(y)dy
\end{equation*}

where $u$ and $v$ satisfy:

\begin{equation*}
    u(x) + v(y) \geq x \cdot y
\end{equation*}

Chartrand et al. \cite{chartrand2009} apply the proposition that $K_D$ has a unique minimizing pair of functions $u$ and $v$ that are convex conjugates of one another (i.e. $u(x) = v^*(x)$ and $v(y) = u^*(y)$). As a result, the derivative of $K_D$ can be found, and thus the authors develop a gradient descent approach to update $u$ and compute the optimal transport map:

\begin{equation*}
    u_{t+1} = u_t - \eta \frac{dK_D(u)}{du}
\end{equation*}
where $\eta$ is the learning rate. For a full derivation, we refer the reader to \cite{chartrand2009}.

\paragraph{Variational Optimal Transport (VOT)}
\label{par:vot}

Similarly to the work of Haker et al. \cite{haker2004}, Kundu et al. \cite{kundu2017} compute the optimal transport map from the Monge formulation of the optimal transport problem. However, the authors modify the objective function in Eq. \eqref{eq:monge-objective} to help guide the solution towards a curl-free mapping. The optimization problem is relaxed further to incorporate the mass-preserving constraint in Eq. \eqref{eq:jacobian-equation}, resulting in the following objective function:

\begin{multline*}
    \psi(f) = \frac{1}{2}\int_X \|f(x)-x\|^2 I_0(x)dx + \frac{\gamma}{2}\int_X \|\nabla \times f(x)\|^2 dx\\
    + \frac{\lambda}{2}\int_X \big(D(x) I_1(f(x)) - I_0(x)\big)^2
    \label{eq:vot-objective}
\end{multline*}
where $\gamma$ and $\lambda$ are regularization coefficients, and $\nabla \times (\cdot)$ is the curl operator. The objective function is non-convex, so Kundu et al. \cite{kundu2017} use a variational optimization technique to solve it. The authors use the Euler-Lagrange equations for the objective function $\psi$ to derive its gradient $\psi_f$, and use Nesterov's accelerated gradient descent method \cite{nesterov1983} to compute the optimal transport map. Moreover, a multi-scale technique is employed to guide the method towards the globally optimum solution.

\subsection{Image Classification}
\label{subsec:exp-classification}

In order to demonstrate the linear separability property of the LOT transform, we tested the SPOT method on five classification tasks: determining the number of Gaussian peaks in an image, distinguishing between animal faces, identifying the shape of galaxies, classifying liver nuclei as cancerous or benign, and recognizing handwritten digits. It should be noted that our aim is not to determine the optimal classification method for each task, but rather to experimentally validate the linear separability theorem proposed in Section \ref{subsec:lot-linear-separability}.

\subsubsection{Synthetic Gaussian Data}
\label{subsubsec:class-experiments-gaussian}

\begin{figure}[t!]
    \centering
    \includegraphics[width=84mm]{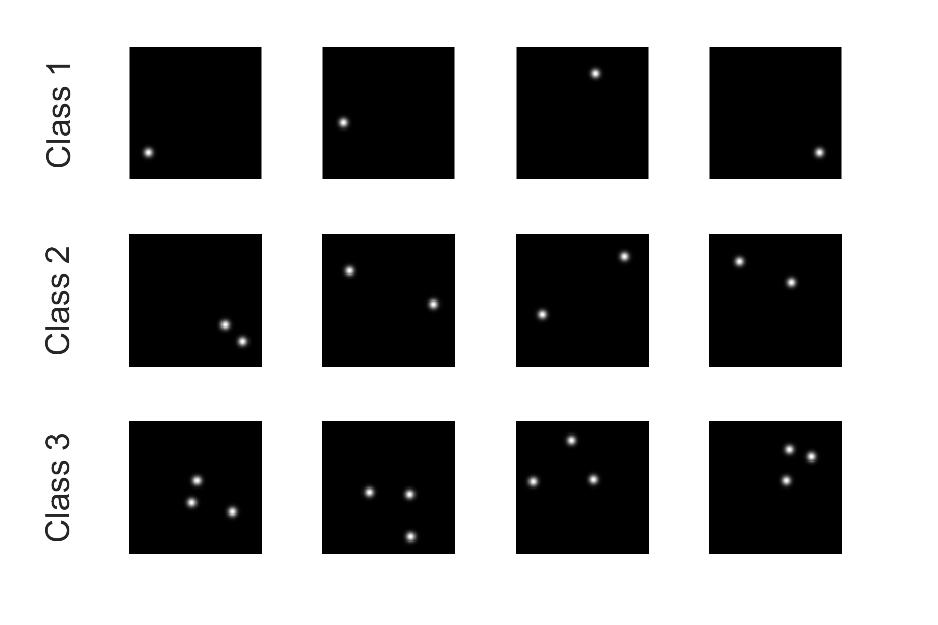}
    \caption{Sample images from each class of the Gaussian dataset. The task is to classify the number of Gaussian peaks in an image.}
    \label{fig:data-gaussian}
\end{figure}

Using the same terminology as Section \ref{subsec:lot-linear-separability}, the ``mother'' densities for the image classes in this dataset are a single 2D Gaussian, a mixture of two 2D Gaussians, and a mixture of three 2D Gaussians. We generated 1000 128${\times}$128px images for each class by randomly translating the mother patterns. To ensure that the sets of images were disjoint, only translations that resulted in non-overlapping Gaussian peaks were selected. Example images from each class are shown in Fig. \ref{fig:data-gaussian}.

After the Gaussian dataset had been generated, we used our SPOT method to compute the transport map -- and by extension, the potential $\phi$ -- between each image $I_i$ and a uniform reference density $I_0$. As the linear separability theorem does not prescribe an optimal classifier, we assessed the classification accuracy of the potential $\phi$ using multiple classifiers: a linear support vector machine (SVM), logistic regression (LR), and penalized Fisher's linear discriminant analysis (PLDA) \cite{wang2011}. Moreover, we employed a ten-fold stratified cross-validation scheme. The dataset was randomly divided into ten subsets, each containing the same proportion of images from each class. Nine subsets were used to train the linear classifier, and the remaining subset was used as the test data. This process was repeated until all ten subsets had been used as the test set. We then computed the mean classification accuracy and standard deviation. For comparison, the classification experiments were conducted on the original images, as well as the potential.

Although classifiers such as SVM and logistic regression are more commonly used than PLDA in the literature, PLDA computes a low-dimensional projection of the data. Therefore, in addition to a quantitative evaluation of the linear separability, PLDA can provide a qualitative visual assessment of the linear separability by projecting the high-dimensional data into two dimensions. Computing PLDA directly on the images or potential involves inverting a matrix of size 128\textsuperscript{2}${\times}$128\textsuperscript{2}. This is computationally expensive, and therefore, prior to classification using PLDA, the size of the input data was reduced using principal component analysis (PCA). In all of our classification experiments, we used the maximum number of principal components (equal to the number of images, in this case) to transform the data. Mathematically, this should have no effect on the classification results, since PCA is linear transformation.

\subsubsection{Animal Faces}
\label{subsubsec:class-experiments-animals}

The LHI-Animal-Faces dataset comprises face images from 20 different classes of animals \cite{si2012}. To complicate the classification of the animals, the categories exhibit a large range of within-class variation, including rotation, posture variation, and different animal sub-types (e.g. deer with and without antlers). In this work, we used a subset of three classes, namely cats, deer, and pandas.

\begin{figure}[t!]
    \centering
    \includegraphics[width=84mm]{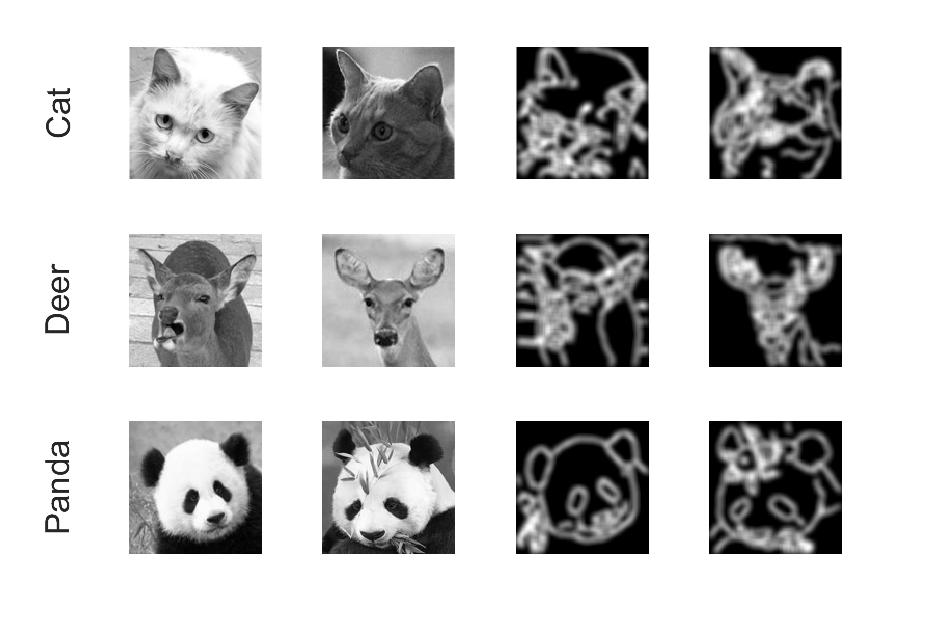}
    \caption{Sample images from the animal faces dataset (left), and the corresponding smoothed edge maps used in the classification experiment described in Section \ref{subsubsec:class-experiments-animals} (right).}
    \label{fig:data-animals}
\end{figure}

In order to better fit the data requirements of the linear separability theorem, the images were pre-processed using the same method as \cite{kolouri2016c}. Firstly, an edge map was generated by applying Canny edge detection to the images. Secondly, the edge maps were smoothed with a Gaussian filter. Example images before and after pre-processing are shown in Fig. \ref{fig:data-animals}.

Following pre-processing, we computed the optimal transport map between each image and a uniform reference image using our SPOT algorithm. Identically to the Gaussian data experiment, we then assessed the classification accuracy of the potential $\phi$ using SVM, logistic regression, and PLDA classifiers with stratified ten-fold cross-validation. For comparison, we also computed the classification accuracy using the pre-processed images.

\subsubsection{Galaxy Shapes}
\label{subsubsec:class-experiments-galaxy}

\begin{figure}[t!]
    \centering
    \includegraphics[width=84mm]{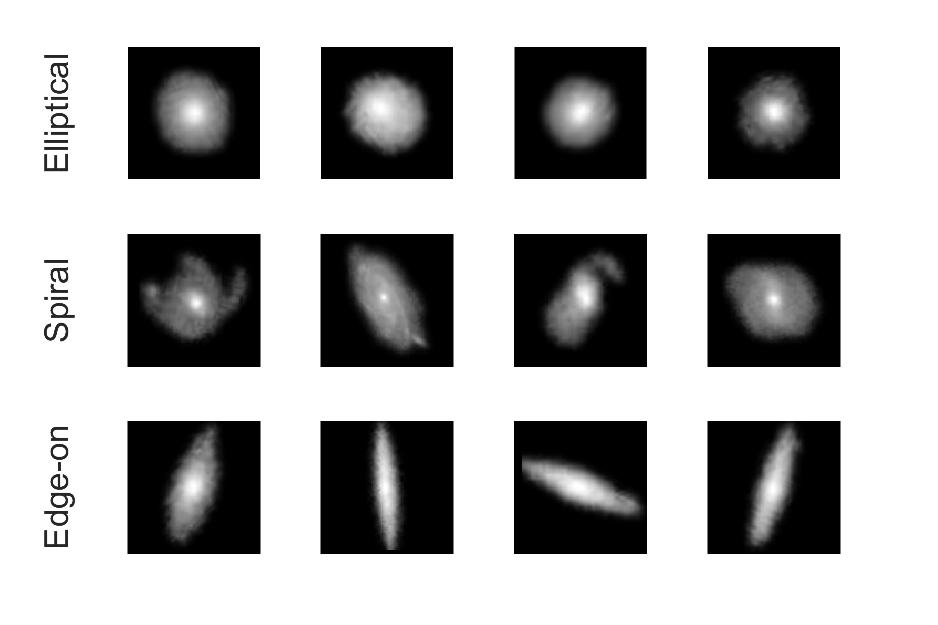}
    \caption{Sample images from each class of the galaxy dataset after pre-processing.}
    \label{fig:data-galaxy}
\end{figure}

In \cite{shamir2009}, Shamir proposed a method that can automatically classify images of elliptical, spiral, and edge-on galaxies. To test the method, the author used a random subset of images from the Galaxy Zoo project \cite{lintott2008}: 225 elliptical galaxies, 224 spiral galaxies, and 75 edge-on galaxies. We used the same subset of Galaxy Zoo data in this work, however, prior to computing the transport maps between the images and a uniform reference, we converted the color images to grayscale, centered the galaxy within each image, and removed background noise by segmenting the galaxy using a thresholding technique. Example pre-processed images are shown in Fig. \ref{fig:data-galaxy}. As before, we compared the classification accuracy of the potential and the original images using three different linear classifiers.

\subsubsection{Liver Nuclei}
\label{subsubsec:class-experiments-liver}

\begin{figure}[t!]
    \centering
    \includegraphics[width=84mm]{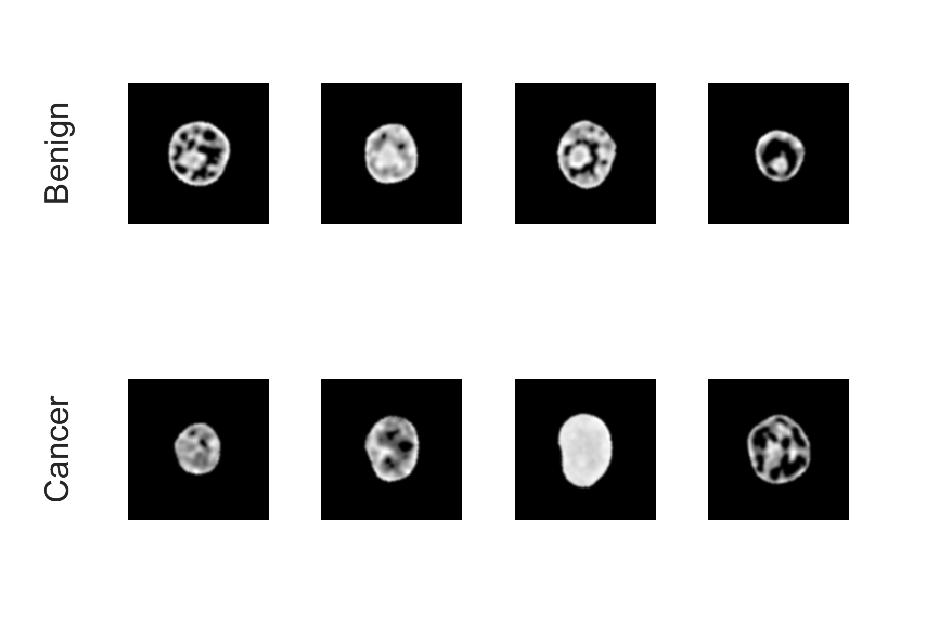}
    \caption{Sample images from both classes of the liver nuclei dataset. Since the patient classes (cancer or healthy) were used as the ground truth labels for the cells, there is no guarantee that all of the nuclei in the ``cancer" class are cancerous.}
    \label{fig:data-liver}
\end{figure}

Another classification task is to classify liver cell nuclei as either fetal-type hepatoblastoma (a type of cancer) or benign. The dataset consists of nuclei images from 17 cancer patients and 26 healthy controls. Each patient has between 60 and 187 nuclei images (mean: 96, standard deviation: 16), resulting in 4145 images in total. Although there is no guarantee that all of the cells from the cancer patients are cancerous, we used the subject label (cancer or benign) and the ground truth for each image. Example images from each class are shown in Fig. \ref{fig:data-liver}.

Unlike the other classification experiments, in which the transport maps were computed using a uniform reference image, for this task, we used a wide Gaussian distribution as the reference. The Gaussian had a similar diameter to the nuclei, resulting a more task-specific reference image. The remainder of the classification experiment was identical to those described in the previous sections.

\subsubsection{MNIST}
\label{subsubsec:class-experiments-mnist}

The final classification task is to assess the linear separation theorem on the MNIST database of handwritten digits. MNIST has become a benchmark dataset for testing machine learning methods, and by using deep neural networks, classification errors of less than 0.25\% have been reported in the literature \cite{wan2013,ciresan2012}. The aim of this work is not to present state-of-the-art classification results on MNIST, but rather, to assess the ability of linear classifiers to distinguish between the classes in potential space compared to image space.

The dataset comprises 60000 training images and 10000 test images, however, to be consistent with our other classification experiments, we used ten-fold cross-validation on the entire set of 70000 images when evaluating the three linear classifiers in image space and potential space. The transport maps (and therefore, the potential) were computed using a uniform reference image $I_0$.
\section{Results}
\label{sec:results}

\subsection{Image Warping Results}
\label{subsec:res-registration}

The optimum SPOT and multi-SPOT parameters for each dataset are presented in Table \ref{tab:parameters}. The table shows that a standard deviation of $\sigma = 1$ universally achieved the lowest mean relative MSE on the training data. Similarly, a gradient descent learning rate of $\eta = 0.01$ was selected at the finest resolution across all datasets. In contrast to the LPBA40 dataset, the optimum values for $\sigma$ and $\eta$ are larger at the coarser scales for the YaleB and ImageNet datasets.

\begin{table}[t!]
    \renewcommand{\arraystretch}{1.3}
    \caption{The optimum parameters for the single-scale and multi-scale versions of our SPOT method, across all three datasets. The multi-SPOT parameters are ordered from the coarsest scale to the finest scale, with $\sigma$ given in units of pixels at the original (finest) image resolution.}
    \label{tab:parameters}
    \centering
    \begin{tabular}{|l|cc|cc|} 
        \hline
        \multirow{2}{4em}{} & \multicolumn{2}{c|}{SPOT} & \multicolumn{2}{c|}{Multi-SPOT}\\
        & $\sigma$ & $\eta$ & $\sigma$ & $\eta$\\
        \hline
        YaleB & 1 & 0.01 & $\{12, 4, 1\}$ & \{1, 0.1, 0.01\}\\
        LPBA40 & 1 & 0.01 & $\{4, 2, 1\}$ & \{0.01, 0.01, 0.01\}\\
        ImageNet & 1 & 0.01 & $\{12, 4, 1\}$ & \{1, 0.1, 0.01\}\\
        \hline
    \end{tabular}
\end{table}

\begin{figure}[t!]
    \centering
        \subfloat[Relative MSE]{\includegraphics[width=84mm]{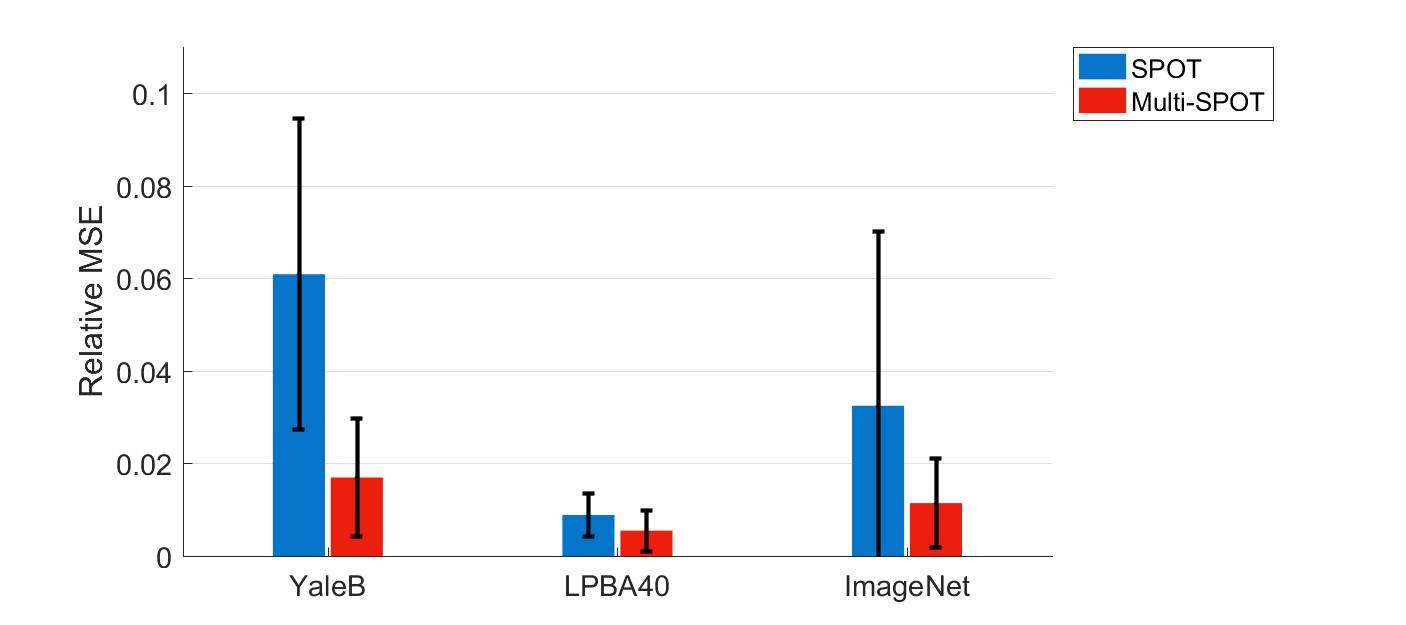}%
        \label{fig:bar-spot-relmse}}
    \hfill
        \subfloat[Mass transported]{\includegraphics[width=84mm]{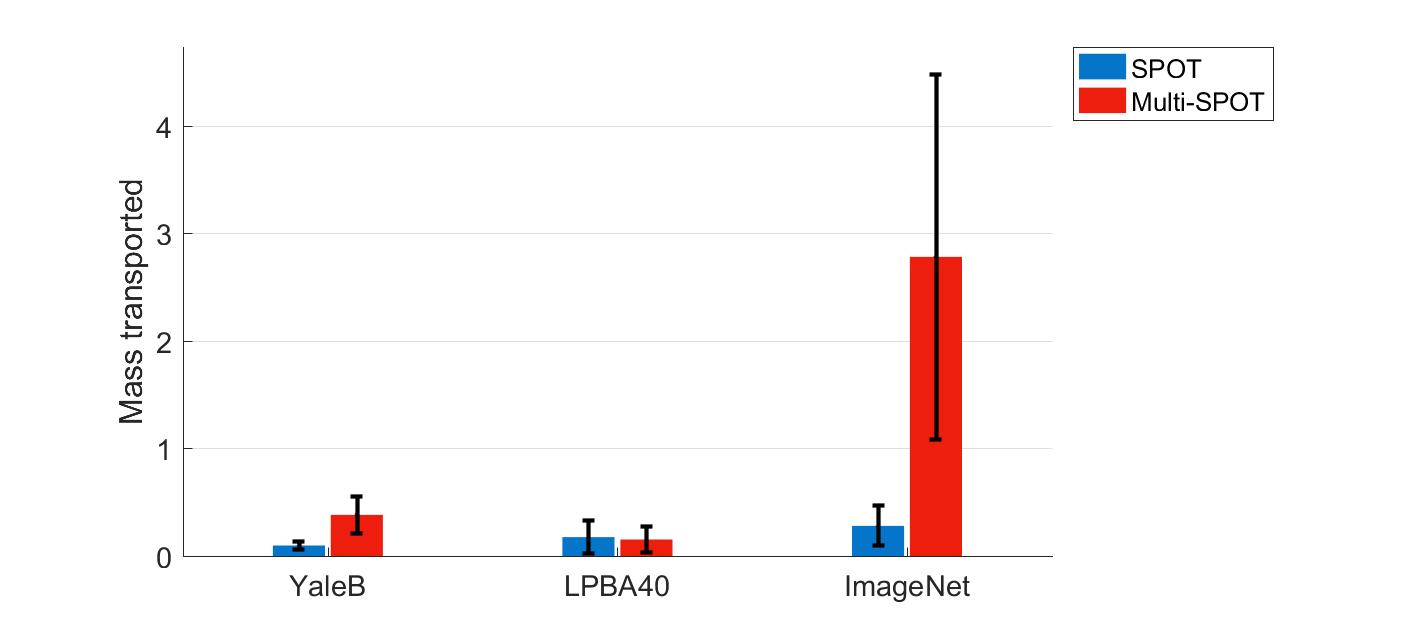}%
        \label{fig:bar-spot-mass}}
    \caption{Mean and standard deviation of the \protect\subref{fig:bar-spot-relmse} relative MSE and \protect\subref{fig:bar-spot-mass} mass transported for the single- and multi-scale SPOT methods on all three datasets. The results for the mean absolute curl have been omitted, since both methods achieved zero curl on all of the datasets.}
    \label{fig:bar-spot}
\end{figure}

Figure \ref{fig:bar-spot} compares the mean and standard deviation of the relative MSE and mass transported for the SPOT and multi-SPOT methods. For all of the datasets tested, multi-SPOT resulted in a statistically lower mean relative MSE than SPOT ($p < 0.01$), and had a smaller standard deviation. Conversely, the SPOT method achieved a significantly lower mean mass transported than multi-SPOT on the YaleB and ImageNet datasets ($p < 0.01$). Although the mean mass transported for multi-SPOT was only slightly lower than the mean mass transported for SPOT on the LPBA40 dataset (0.15 and 0.18, respectively), the two means are significantly different ($p < 0.01$). The final evaluation measure, mean absolute curl, has been omitted from Fig. \ref{fig:bar-spot} because both methods achieved zero curl for all three datasets. This is because the curl of a potential is zero. 

\begin{figure*}[t!]
    \centering
        \subfloat[YaleB]{\includegraphics[width=129mm]{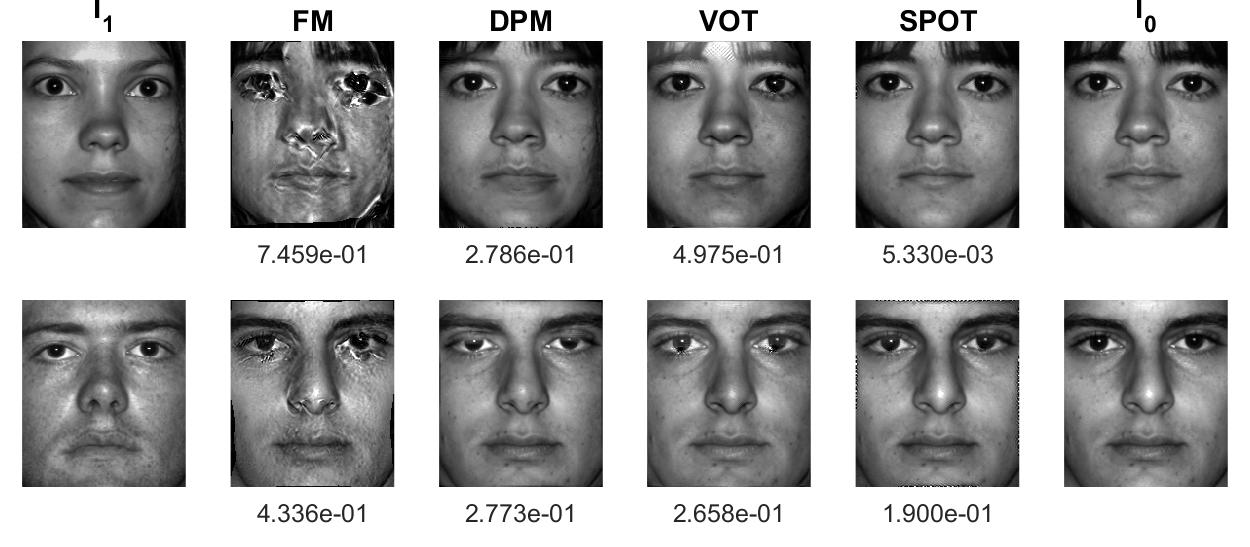}%
        \label{fig:results-yaleb-spot}}
    \hfill
        \subfloat[LPBA40]{\includegraphics[width=129mm]{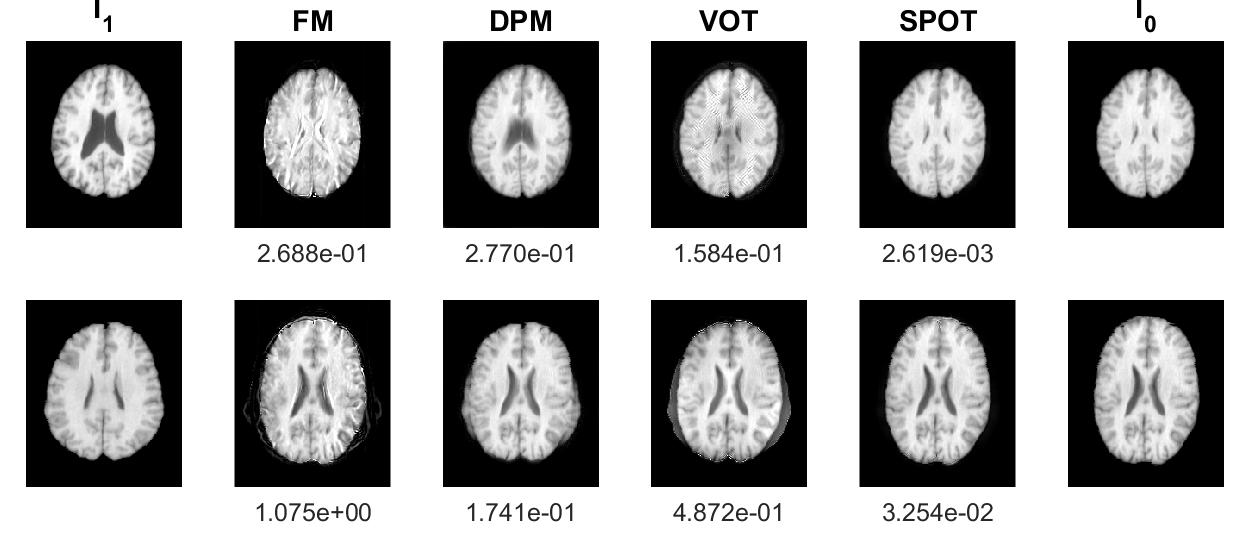}%
        \label{fig:results-lpba40-spot}}
    \hfill
        \subfloat[ImageNet]{\includegraphics[width=129mm]{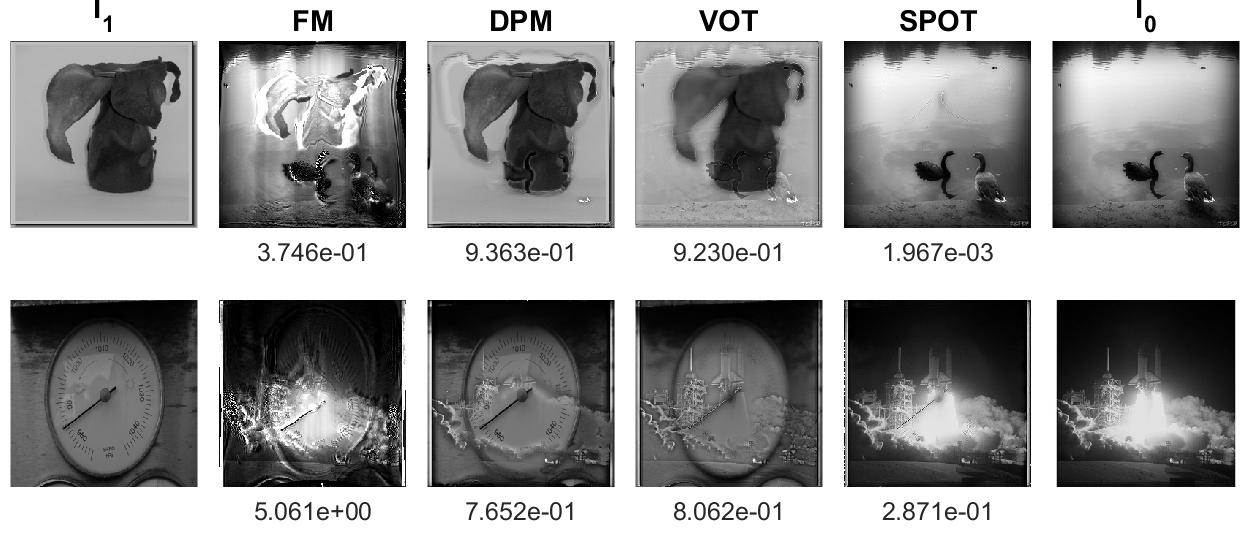}%
        \label{fig:results-imagenet-spot}}
    \caption{Example results for the single-scale optimal transport methods on \protect\subref{fig:results-yaleb-spot} the YaleB dataset, \protect\subref{fig:results-lpba40-spot} the LPBA40 dataset, and \protect\subref{fig:results-imagenet-spot} the ImageNet dataset. Columns 2-5 show the warped input image $D_f(x)I_1(f(x))$ for each method after attempting to map $I_1$ to $I_0$. The relative MSE of each result is displayed below the image. The top and bottom rows of each subfigure show the mapping that resulted in the lowest and highest relative MSE, respectively, for the SPOT method.}
    \label{fig:results-spot}
\end{figure*}

Figure \ref{fig:results-spot} shows example results for the single-scale optimal transport methods (FM, DPM, VOT and SPOT) on all three datasets. The central columns show the warped input image $D_f(x)I_1(f(x))$ for each method, after attempting to map $I_1$ to $I_0$. The relative MSE of each result is displayed below the warped image. The top and bottom rows of each subfigure show the mapping that resulted in the lowest and highest relative MSE, respectively, for the SPOT method. In all of the examples in Fig. \ref{fig:results-spot}, the relative MSE for the SPOT method was lower than the corresponding relative MSEs for the other single-scale optimal transport methods. Moreover, the lowest relative MSE for the SPOT method was two orders of magnitude smaller than the corresponding relative MSEs of the other methods.


\begin{figure*}[t!]
    \centering
        \subfloat[YaleB]{\includegraphics[width=84mm]{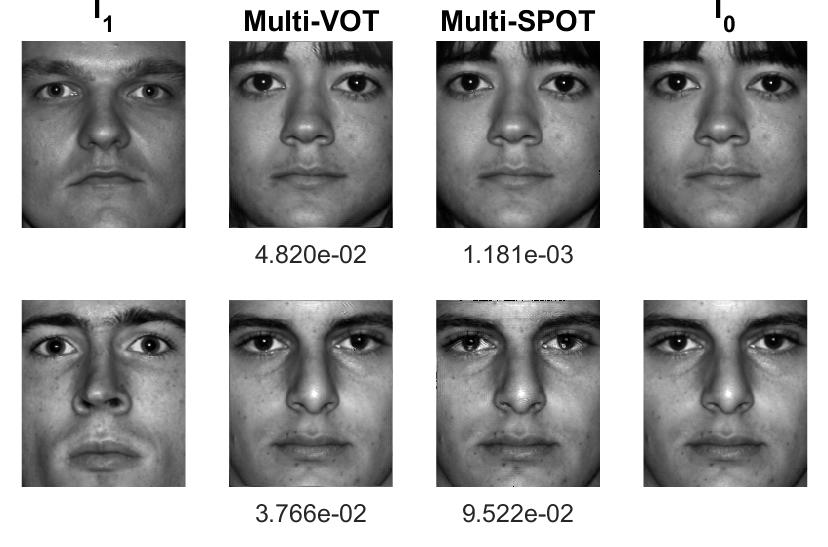}%
        \label{fig:results-yaleb-multispot}}
    \hfill
        \subfloat[LPBA40]{\includegraphics[width=84mm]{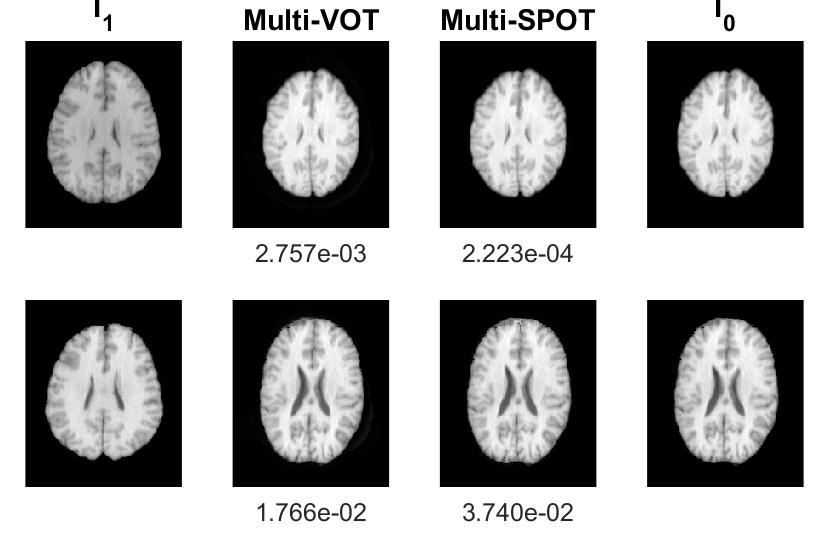}%
        \label{fig:results-lpba40-multispot}}
    \hfill
        \subfloat[ImageNet]{\includegraphics[width=84mm]{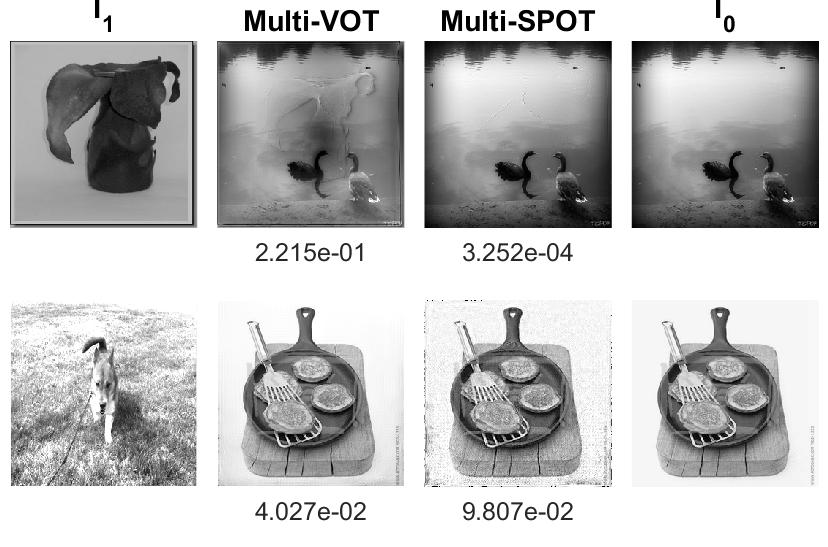}%
        \label{fig:results-imagenet-multispot}}
    \caption{Example results for the multi-scale optimal transport methods on \protect\subref{fig:results-yaleb-multispot} the YaleB dataset, \protect\subref{fig:results-lpba40-multispot} the LPBA40 dataset, and \protect\subref{fig:results-imagenet-multispot} the ImageNet dataset. Columns 2-3 show the warped input image $D_f(x)I_1(f(x))$ for each method after attempting to map $I_1$ to $I_0$. The relative MSE of each result is displayed below the image. The top and bottom rows of each subfigure show the mapping that resulted in the lowest and highest relative MSE, respectively, for the multi-SPOT method.}
    \label{fig:results-multispot}
\end{figure*}

Analogously to Fig. \ref{fig:results-spot}, Fig. \ref{fig:results-multispot} shows example results for the multi-scale optimal transport methods (multi-VOT and multi-SPOT) on all three datasets. The second and third columns show the warped input image $D_f(x)I_1(f(x))$ for each method, after attempting to map $I_1$ to $I_0$. The relative MSE of each result is displayed below the warped image. The top and bottom rows of each subfigure show the mapping that resulted in the lowest and highest relative MSE, respectively, for the multi-SPOT method. Similarly to the single-scale optimal transport results, in all three datasets, the lowest relative MSE for the multi-SPOT method was an order of magnitude lower than the corresponding relative MSE for the multi-VOT method. However, the cases that produced the highest relative MSE for the multi-SPOT method resulted in a lower relative MSE in the multi-VOT method.

\begin{figure}[t!]
    \centering
        \subfloat[Relative MSE]{\includegraphics[width=84mm]{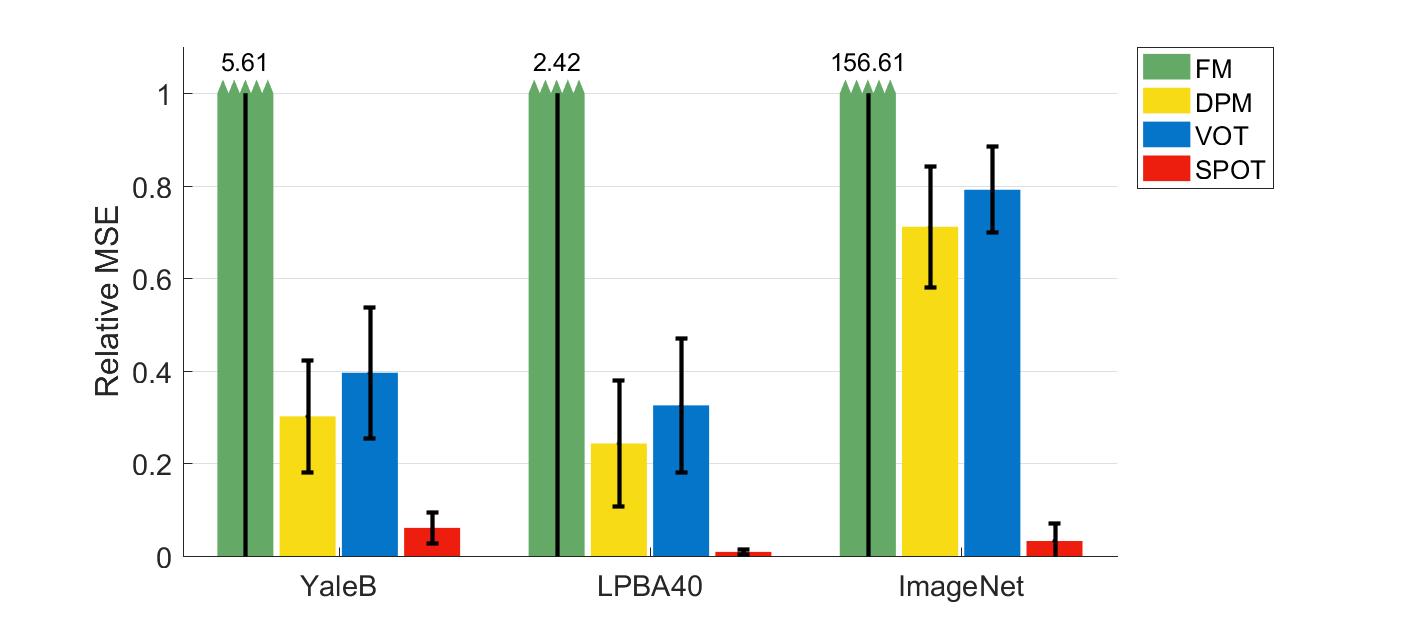}%
        \label{fig:bar-single-relmse}}
    \hfill
        \subfloat[Mass transported]{\includegraphics[width=84mm]{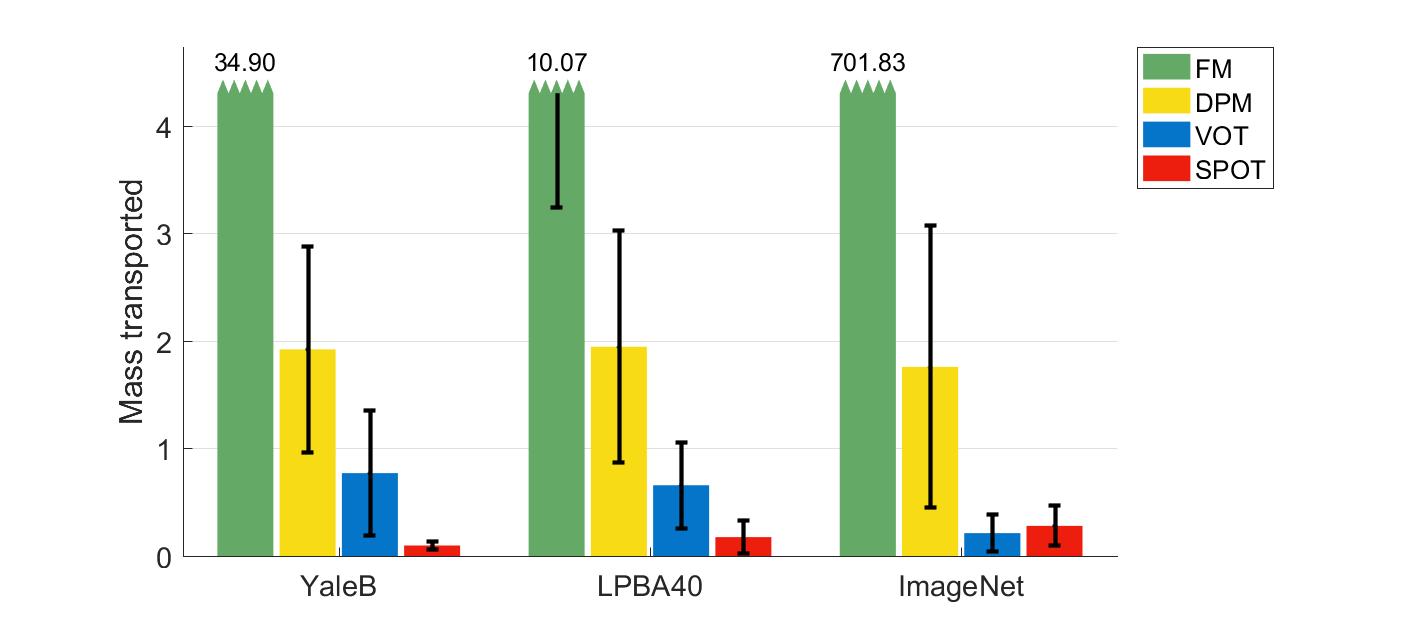}%
        \label{fig:bar-single-mass}}
    \hfill
        \subfloat[Mean absolute curl]{\includegraphics[width=84mm]{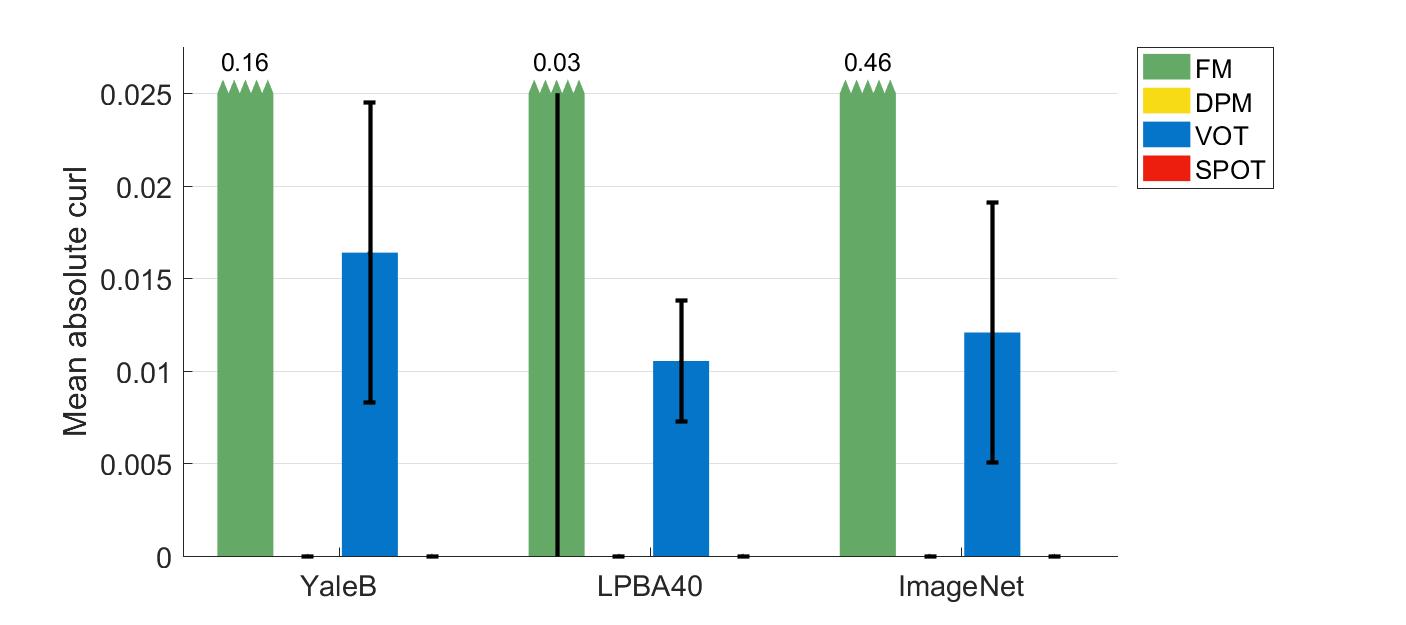}%
        \label{fig:bar-single-mac}}
    \caption{Mean and standard deviation of the \protect\subref{fig:bar-single-relmse} relative MSE, \protect\subref{fig:bar-single-mass} mass transported, and \protect\subref{fig:bar-single-mac} mean absolute curl, for the single-scale optimal transport methods on all three datasets. For clarity, large bars have been truncated, and the true mean value of the error measure is displayed above the bar.}
    \label{fig:bar-single}
\end{figure}

The mean and standard deviation of the error measures for the single-scale optimal transport methods (FM, DPM, VOT, and SPOT) are presented in Fig. \ref{fig:bar-single}. For clarity, large bars have been truncated, and the true mean value of the error measure is displayed above the bar. Across all three datasets, SPOT resulted in a significantly lower mean relative MSE that the other single-scale optimal transport methods ($p < 0.01$, corrected for multiple comparisons). The standard deviation of the relative MSE was also smaller for SPOT compared to the other methods. A similar trend is observed for the mass transported in Fig. \ref{fig:bar-single-mass}, except that the VOT method resulted in a significantly lower mean mass transported than SPOT ($p < 0.01$) for the ImageNet dataset. Since the SPOT and DPM methods are based on Brenier's theorem and derive the transport maps from a potential, the mean and standard deviation of the mean absolute curl were universally zero for those two methods .

\begin{figure}[t!]
    \centering
        \subfloat[Relative MSE]{\includegraphics[width=84mm]{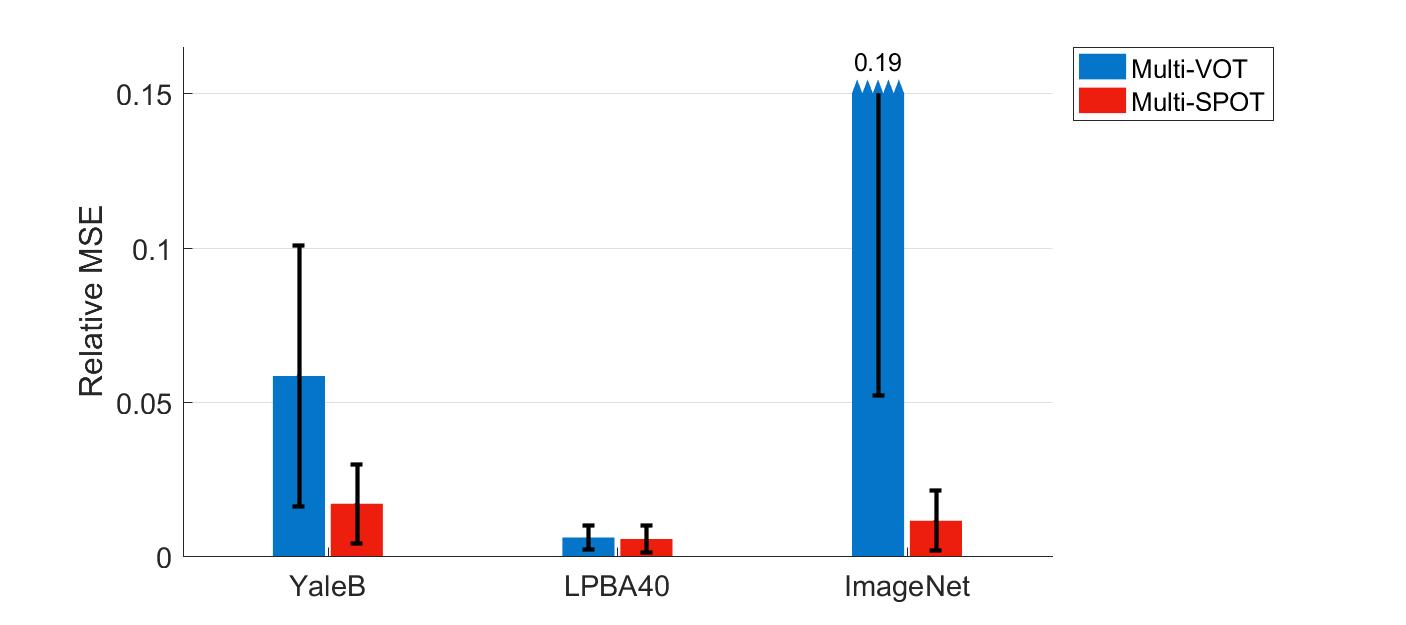}%
        \label{fig:bar-multi-relmse}}
    \hfill
        \subfloat[Mass transported]{\includegraphics[width=84mm]{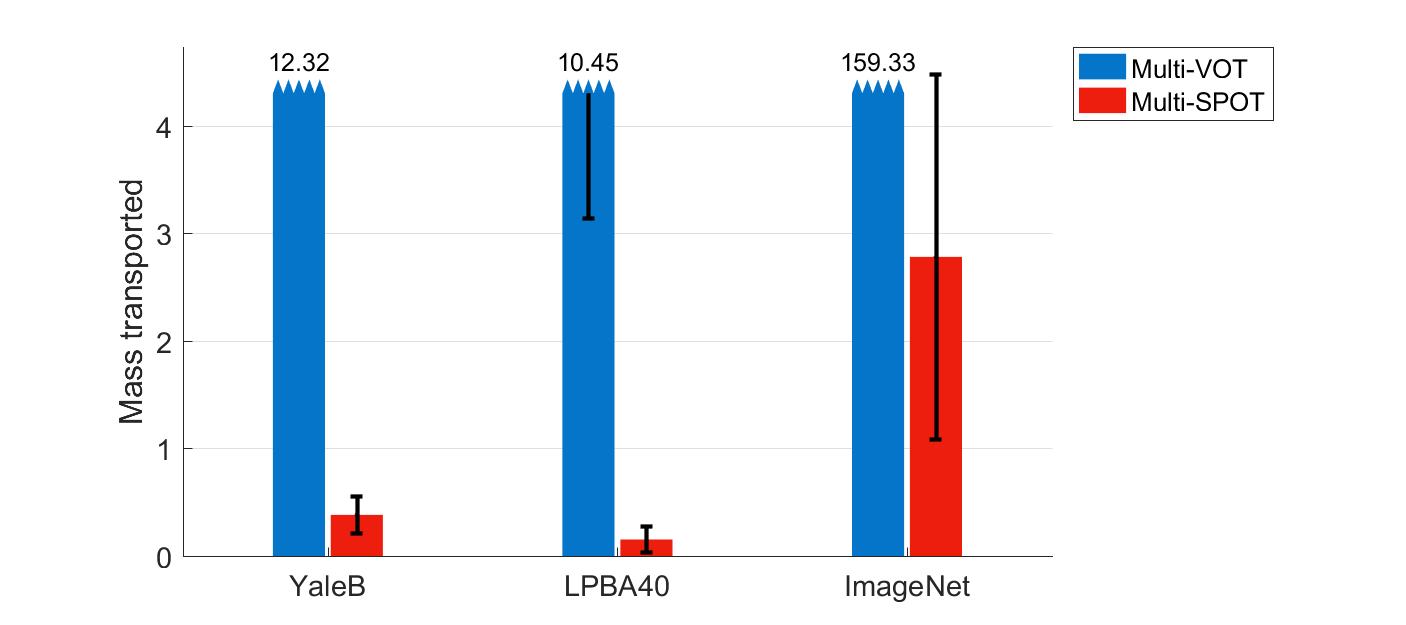}%
        \label{fig:bar-multi-mass}}
    \hfill
        \subfloat[Mean absolute curl]{\includegraphics[width=84mm]{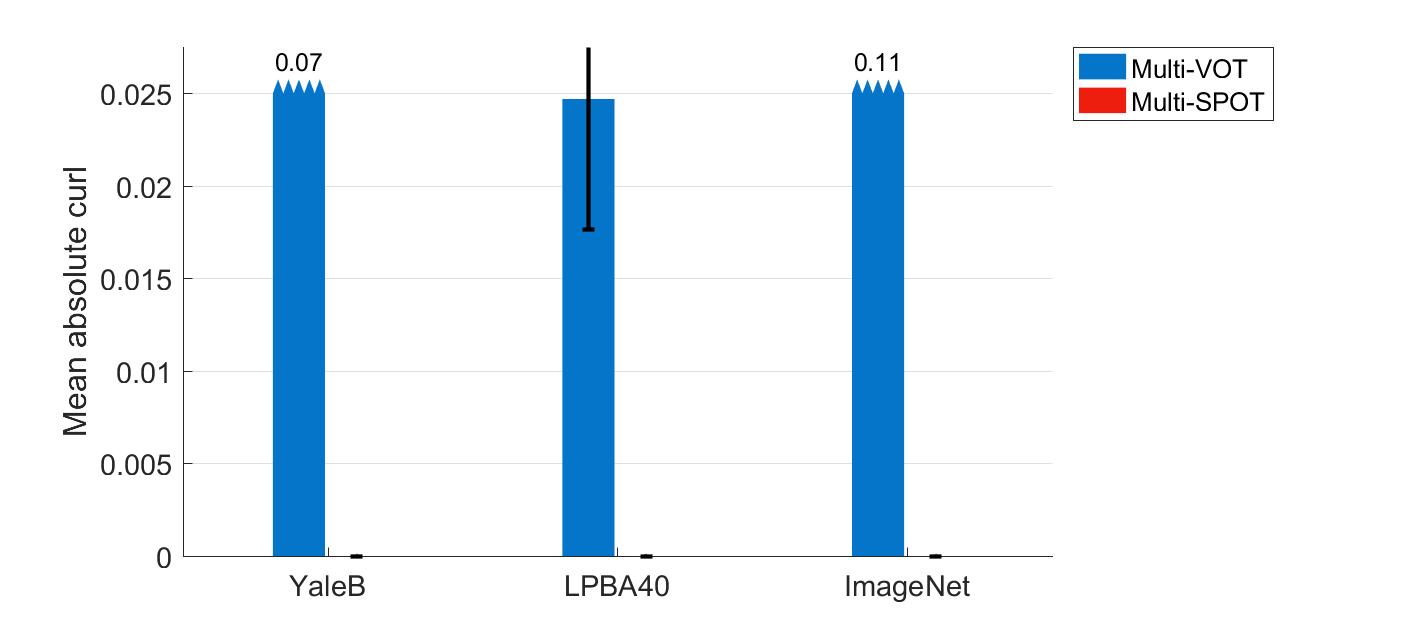}%
        \label{fig:bar-multi-mac}}
    \caption{Mean and standard deviation of the \protect\subref{fig:bar-multi-relmse} relative MSE, \protect\subref{fig:bar-multi-mass} mass transported, and \protect\subref{fig:bar-multi-mac} mean absolute curl, for the multi-scale optimal transport methods on all three datasets. For clarity, large bars have been truncated, and the true mean value is displayed above the bar. It should also be noted that the relative MSE in \protect\subref{fig:bar-multi-relmse} is shown at a smaller scale than Figure \protect\ref{fig:bar-single-relmse}.}
    \label{fig:bar-multi}
\end{figure}

Figure \ref{fig:bar-multi} shows the mean and standard deviation of the error measures for the multi-scale optimal transport methods (multi-VOT and multi-SPOT). Identically to Fig. \ref{fig:bar-single}, large bars have been truncated, and the true mean value of the error measure is displayed above the bar. For convenience of comparison, the mass transported and mean absolute curl are shown at the same scale as Fig. \ref{fig:bar-single}, however, the relative MSE is shown at a smaller scale to reflect the lower relative MSEs for the multi-scale optimal transport methods. For all error measures and datasets, multi-SPOT resulted in a significantly lower mean value than multi-VOT ($p < 0.01$). Moreover, multi-SPOT achieved a smaller standard deviation than multi-VOT for all error measures and datasets, except for relative MSE on the LPBA40 dataset.

\subsection{Classification Results}
\label{subsec:res-classification}

\begin{figure}[t!]
    \centering
        \subfloat[Image space]{\includegraphics[width=84mm]{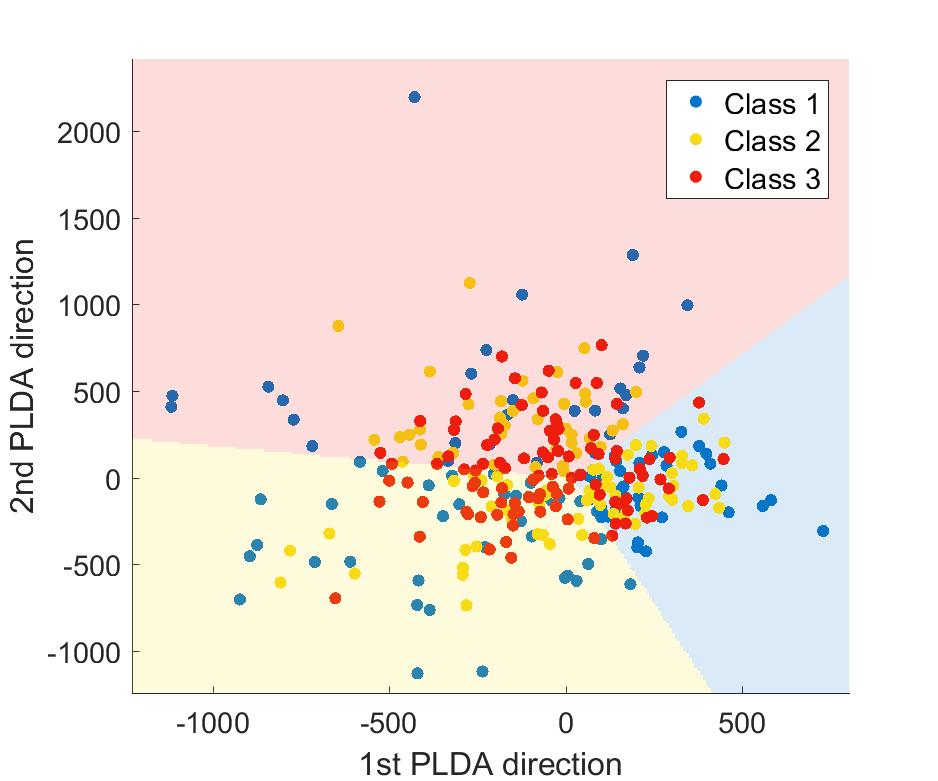}%
        \label{fig:classify-gaussian-img}}
    \hfill
        \subfloat[Potential space]{\includegraphics[width=84mm]{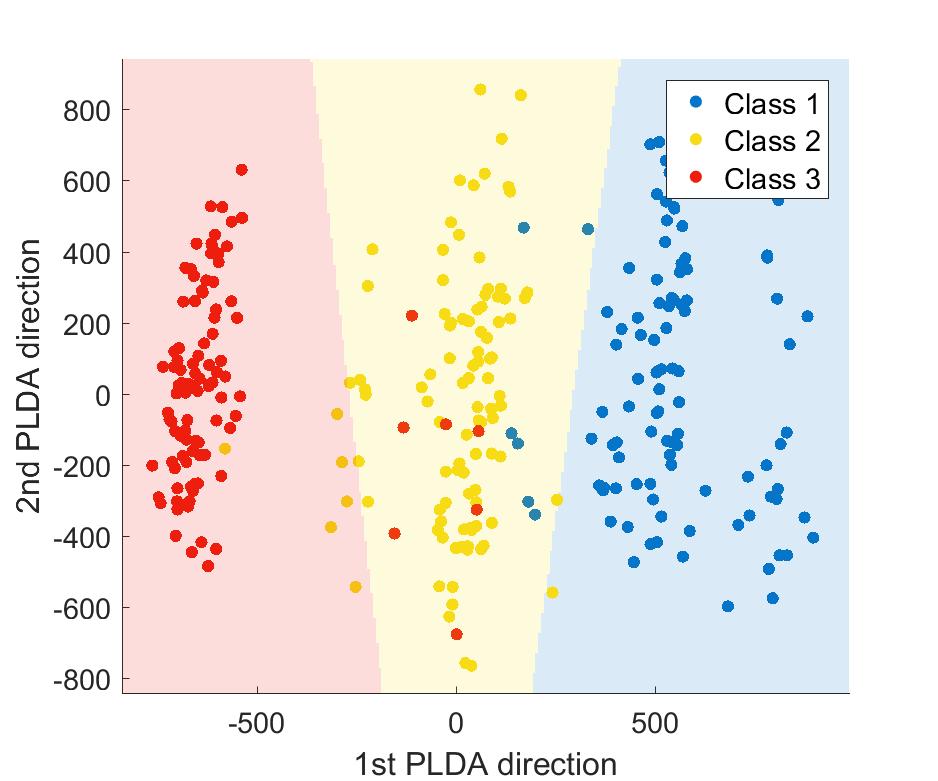}%
        \label{fig:classify-gaussian-phi}}
    \caption{Two-dimensional PLDA projection of \protect\subref{fig:classify-gaussian-img} the original Gaussian image data, and \protect\subref{fig:classify-gaussian-phi} its potential $\phi$. In image space, the classes are indistinguishable, whereas in potential space they are almost perfectly linearly separable. The results shown here correspond to a single set of test data from the ten-fold cross validation.}
    \label{fig:classify-gaussian}
\end{figure}

The mean and standard deviation of the classification accuracy from the five classification tasks are presented in Table \ref{tab:classify-results}. For the task in which the classifiers have to determine the number of Gaussian peaks in an image, the mean classification accuracy of the original image data was only marginally higher than chance for all three classifiers. In contrast, when classifying the data using the potential $\phi$, the mean accuracy was around 90\%. This result is illustrated in Fig. \ref{fig:classify-gaussian}, where the classes are indistinguishable in the two-dimensional PLDA projection of the image data, but are almost perfectly linearly separable in the potential space.

Although the increase in classification accuracy between image and potential space is smaller for the animal faces dataset than the Gaussians dataset, a higher accuracy was achieved across all three classifiers when using the potential. Interestingly, the accuracy of the logistic regression on the animal faces image data was considerably lower than the accuracy obtained using a SVM or PLDA on the same data. In contrast, all three linear classifiers resulted in a very similar classification accuracy ($\sim$85\%) for the potential derived from the animal faces data.

For the galaxy and liver nuclei datasets, classification on the potential only resulted in a higher mean accuracy than image space for two out of the three classifiers. However, the accuracies on image space and potential space were very close for the cases where classification on image space achieved a higher accuracy. Finally, for the MNIST dataset, classification on potential space resulted in a lower accuracy than image space using the PLDA classifier. The results for the SVM and logistic regression classifiers were almost identical in both image space and potential space, with image space achieving a marginally smaller standard deviation.

\begin{table*}[t!]
    \renewcommand{\arraystretch}{1.3}
    \caption{Mean $\pm$ standard deviation classification accuracy on the images and potential $\phi$ using a linear SVM, logistic regression, and PLDA. The best results for each classifier are highlighted in bold font.}
    \label{tab:classify-results}
    \centering
    \begin{tabular}{|l|c|c|c|c|c|c|} 
        \hline
        \multirow{2}{4em}{Dataset} & \multicolumn{2}{c|}{SVM} & \multicolumn{2}{c|}{Logistic regression} & \multicolumn{2}{c|}{PLDA}\\
        \cline{2-7}
        & Image & $\phi$ & Image & $\phi$ & Image & $\phi$\\
        \hline
        Gaussians & 0.348${\pm}$0.011 & \textbf{0.882${\pm}$0.030} & 0.347${\pm}$0.020 & \textbf{0.895${\pm}$0.013} & 0.357${\pm}$0.024 & \textbf{0.930${\pm}$0.012}\\
        Animal faces & 0.714${\pm}$0.078 & \textbf{0.851${\pm}$0.058} & 0.423${\pm}$0.007 & \textbf{0.851${\pm}$0.064} & 0.784${\pm}$0.071 & \textbf{0.830${\pm}$0.039}\\
        Galaxies & 0.636${\pm}$0.048 & \textbf{0.773${\pm}$0.034} & 0.698${\pm}$0.054 & \textbf{0.810${\pm}$0.034} & 0.\textbf{805${\pm}$0.057} & 0.788${\pm}$0.058\\
        Liver nuclei & \textbf{0.762${\pm}$0.019} & 0.759${\pm}$0.020 & 0.769${\pm}$0.019 & \textbf{0.771${\pm}$0.017} & 0.732${\pm}$0.029 & \textbf{0.763${\pm}$0.030}\\
        MNIST & \textbf{0.917${\pm}$0.003} & 0.917${\pm}$0.004 & \textbf{0.917${\pm}$0.003} & 0.917${\pm}$0.004 & \textbf{0.810${\pm}$0.003} & 0.737${\pm}$0.005\\
        \hline
    \end{tabular}
\end{table*}
\section{Discussion}
\label{sec:discussion}

In the first set of experiments, we compared our numerical implementation of the LOT transform, based on the solution of the Monge-Amp\`{e}re equation, to existing optimal transport methods from the literature. We showed that our SPOT method is able to provide a curl-free mapping between two images, which results in a lower error than other methods. This is highlighted by the high-quality image matches in Figs. \ref{fig:results-spot} and \ref{fig:results-multispot}, as well as the plots of mean MSE in Figs. \ref{fig:bar-single} and \ref{fig:bar-multi}. For all four methods tested in Section \ref{subsec:res-registration}, the lowest mean relative MSE was obtained on the LPBA40 dataset. This could be due to the relatively small deformations and intensity changes required to match the brain images compared to the face images and ImageNet data. This is also reflected in the optimum multi-SPOT parameters presented in Table \ref{tab:parameters}. Wider Gaussian basis functions are able to capture larger deformations, and larger $\sigma$ values were selected for the coarsest scales of the YaleB and ImageNet datasets compared to the LPBA40 dataset. 

Despite achieving a lower mean mass transported than the multi-VOT method, multi-SPOT resulted in a larger mean mass transported than its single-scale equivalent. This is possibly because our cost function in Eq. \eqref{eq:spot-objective} does not directly minimize the mass transported, but rather, the MSE. As a result, the multi-scale approach to SPOT achieves a significantly lower mean MSE than the single-scale approach, at the expense of a smaller mean mass-transported. However, the single- and multi-scale versions of our algorithm resulted in a considerably lower mass transported than the other methods, suggesting that the map obtained by our numerical implementation is closer to the true optimal transport map.

The aim of the classification experiments described in Section \ref{subsec:exp-classification} was to demonstrate the LOT linear separability theorem stated in Section \ref{subsec:lot-linear-separability}. Using our synthetic Gaussian dataset, specifically designed to match the mathematical criteria of the linear separability theorem, we showed that the data are linearly inseparable in image space, and almost perfectly separable in transport space. A visual inspection of the few misclassified results in Fig. \ref{fig:classify-gaussian-phi} reveals that these images contain Gaussian peaks close to the edge of the image. Consequently, slight errors in accurately computing the potential at the boundary could have affected the classification result.

\begin{figure}[t!]
    \centering
        \subfloat[]{\includegraphics[width=39mm]{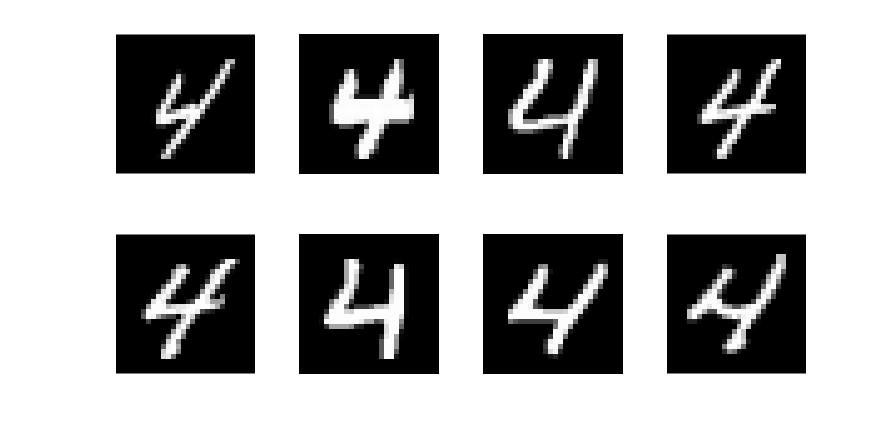}%
        \label{fig:misclass-mnist-44}}
    \hfill
        \subfloat[]{\includegraphics[width=39mm]{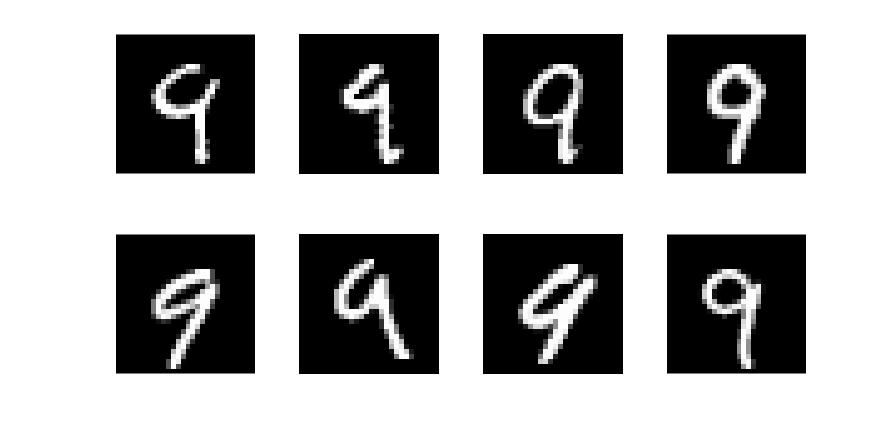}%
        \label{fig:misclass-mnist-99}}
    \hfill
        \subfloat[]{\includegraphics[width=39mm]{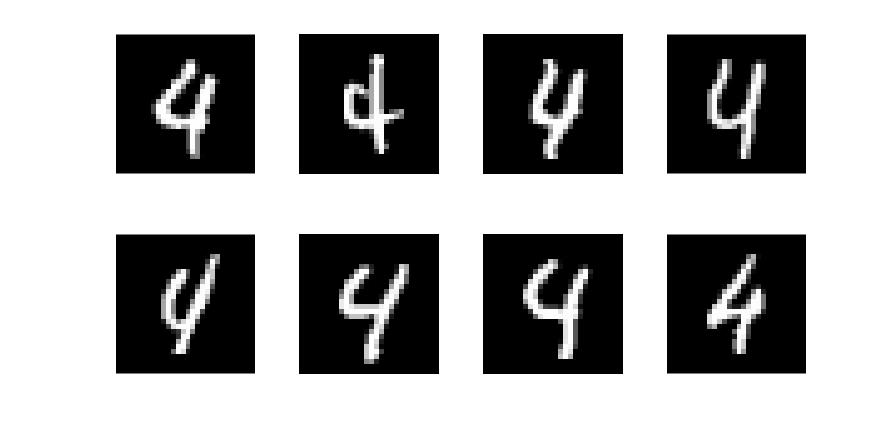}%
        \label{fig:misclass-mnist-49}}
    \caption{Images from the MNIST dataset that were \protect\subref{fig:misclass-mnist-44} correctly identified as the digit ``4'' by the PLDA classfier in both spaces, \protect\subref{fig:misclass-mnist-99} correctly identified as the digit ``9'' by the PLDA classifier in both spaces, and \protect\subref{fig:misclass-mnist-49} correctly classified as 4's by the PLDA classfier in image space, but misclassified as 9's in potential space.}
    \label{fig:misclass-mnist}
\end{figure}

Other than the PLDA classifier on the MNIST dataset, classification on the potential achieved a similar, if not higher, accuracy than classification in image space using the same linear classifier. This suggests that classification in LOT space can be beneficial on real data as well as synthetic examples. Moreover, the relatively lower classification accuracy of the PLDA classifier on the MNIST dataset could be due to the handwritten digits not fulfilling the mathematical requirements of the linear separability theorem. For example, the confusion matrix of classification results indicates that images containing the digit ``4'' were more commonly misclassified as ``9'' in potential space compared to image space. Example images of 4's and 9's that were correctly classified by the PLDA classifier in image space and potential space are shown in Figs. \ref{fig:misclass-mnist-44} and \ref{fig:misclass-mnist-99}. Images that were correctly identified as 4's in image space, but incorrectly classified as 9's in potential space, are shown in Fig. \ref{fig:misclass-mnist-49}. In comparison to the correctly classified 4's, the misclassified 4's appear to be less angular, with the top part of the digit forming a loop similar to several of the correctly identified 9's. This suggests that certain classes in the MNIST dataset are not completely disjoint, and thus do not meet the requirements of the linear separability theorem outlined in Section \ref{subsec:lot-linear-separability}.

\subsection{Summary and Conclusions}
\label{subsec:conclusion}


In this work we have presented a pattern theoretic approach for modeling and inference from image data based on the mathematics of optimal transport. Inspired by the work described in \cite{park2017,kolouri2016c}, we have framed the linear optimal transport approach developed in \cite{wang2013,basu2014,kolouri2016b} as a nonlinear signal transform, and described some of its mathematical properties. The transform involves computing a transport map between a given image and a chosen reference. Hence -- and in contrast to linear signal transforms such as the Fourier and Wavelet transforms -- the LOT transform allows for encoding of pixel displacements, and thus allows for the comparison of pixel intensities at non-fixed image coordinates.

It should be noted that the computational results shown here relied on a numerical implementation of the underlying (Monge) OT problem, where the transport map was modeled directly as the gradient of a convex potential function $\phi$. Furthermore, the potential function was itself modeled as a linear sum of smooth basis functions. We showed numerical comparisons to other numerical (Monge) OT approaches in order to verify the accuracy of the computed solutions. We clarify, however, that the signal LOT transformation described here could make use of any numerical solver for actual implementation.


We also note that the nature of the numerical implementation we described here is, in a way, incompatible with the signal transformation properties (e.g. translation) shown earlier. This is because have have used a numerical PDE approach which requires both source and target images to be fixed grids in the square domain $[0,1]^2$. This could explain in part some loss in the ability of the numerical transformation method to make signal classes more linear separable, though the numerical classification results suggest at times the signal transformation framework can be valuable in making the pattern recognition problem simpler to solve. 

We believe the contributions presented here could enhance the theoretical foundations of the emerging transport and other Lagrangian signal transformation methods that have been recently been employed in a variety of applications including modeling brain anatomy \cite{kundu2017} and cell morphology \cite{basu2014}, cancer detection \cite{ozolek2014,tosun2015,hanna2017}, communications \cite{park2018}, modeling turbulence in imagine experiments \cite{nichols2017}, inverse problems \cite{kolouri2015,cattell2017}, and other applications including machine learning and applied statistics \cite{kolouri2017}.

\section{Acknowledgements}
The authors gratefully acknowledge funding from the National Science Foundation (CCF 1421502) and the National Institutes of Health R01 (GM090033) in contributing to this work.

\appendix
\section{Proof of the Translation Property}
\label{app:proof-translation}

Consider two positive density functions $I_0$ and $I_1$ defined on measure spaces $X \subseteq \RR^d$ and $Y \subseteq \RR^d$, respectively. We assume that the total mass associated with $X$ is equal to the total mass associated with $Y$, such that:

\begin{equation}
    \int_{X}I_0(x)dx = \int_{Y}I_1(y)dy
\end{equation}
The optimal transport problem is to find a the least cost transformation that is also a mass-preserving map $f_1 : X \rightarrow Y$ that maps one density to the other:

\begin{equation}
    \int_{A} I_0(x)dx = \int_{f_1(A)} I_1(y)dy \quad \forall A \subset X
    \label{eq:translation-I1-map}
\end{equation}
If we consider $f_1$ to be the gradient of a convex potential $\phi_1$, such that $f_1(x) = \nabla (\frac{1}{2}|x|^2 - \mu^T x)$, then by Brenier's theorem, there exists a unique (up to a constant) transport map $f_1$ that will satisfy the equation above. Now let $I_\mu$ represent a translation of $I_1$ by $\mu$, such that $I_\mu(y) = I_1(y-\mu)$. In the same manner as above, we can solve for $f_\mu : X \rightarrow Y$:

\begin{equation}
    \begin{aligned}
        \int_{A} I_0(x)dx &= \int_{f_\mu(A)} I_\mu(y)dy\\
        &= \int_{f_\mu(A)} I_1(y-\mu)dy
    \end{aligned}
    \label{eq:translation-Imu-map}
\end{equation}
By substituting $u = y - \mu$ into Eq. \eqref{eq:translation-Imu-map} via the change of variables theorem, and equating Eqs. \eqref{eq:translation-I1-map} and \eqref{eq:translation-Imu-map} we get:

\begin{equation}
    \int_{f_1(A)} I_1(y)dy = \int_{f_\mu(A)-\mu} I_1(u)du
\end{equation}
Since the integrals are equal, we have $f_\mu(x) = f_1(x) + \mu$. If we define the transform of density $I$ with respect to the reference density $I_0$ as $\hat{I} = (f(x) - x) \sqrt{I_0(x)}$, then:

\begin{equation}
    \begin{aligned}
        \hat{I}_\mu &= (f_\mu(x) - x) \sqrt{I_0(x)}\\
        &= (f_1(x) - x + \mu) \sqrt{I_0(x)}\\
        &= \hat{I}_1(x) + \mu \sqrt{I_0(x)}
    \end{aligned}
\end{equation}
\section{Proof of the Scaling Property}
\label{app:proof-scaling}

Consider two positive density functions $I_0$ and $I_1$ defined on measure spaces $X \subseteq \RR^d$ and $Y \subseteq \RR^d$, respectively. We assume that the total mass associated with $X$ is equal to the total mass associated with $Y$, such that:

\begin{equation}
    \int_{X}I_0(x)dx = \int_{Y}I_1(y)dy
\end{equation}
The optimal transport problem is to find a mass-preserving map $f_1 : X \rightarrow Y$ that maps one density to the other:

\begin{equation}
    \int_{A} I_0(x)dx = \int_{f_1(A)} I_1(y)dy \quad \forall A \subset X
    \label{eq:scaling-I1-map}
\end{equation}
If we consider $f_1$ to be the gradient of a convex potential $\phi_1$, such that $f_1 = \nabla(\frac{\sigma}{2}|x|^2)$, then by Brenier's theorem, there exists a unique transport map $f_1$ that will satisfy the equation above. Now let $I_\sigma$ represent a scaling of $I_1$ by $\sigma$, such that $I_\sigma(y) = \sigma I_1(\sigma y)$. In the same manner as above, we can solve for $f_\sigma : X \rightarrow Y$:

\begin{equation}
    \begin{aligned}
        \int_{A} I_0(x)dx &= \int_{f_\sigma(A)} I_\sigma(y)dy\\
        &= \int_{f_\sigma(A)} \sigma I_1(\sigma y)dy
    \end{aligned}
    \label{eq:scaling-Imu-map}
\end{equation}
By substituting $u = \sigma y$ into Eq. \eqref{eq:scaling-Imu-map} via the change of variables theorem, and equating Eqs. \eqref{eq:scaling-I1-map} and \eqref{eq:scaling-Imu-map} we get:

\begin{equation}
    \int_{f_1(A)} I_1(y)dy = \int_{\sigma f_\sigma(A)} I_1(u)du
\end{equation}
Since the integrals are equal, we have $f_\sigma(x) = \frac{f_1(x)}{\sigma}$. If we define the transform of density $I$ with respect to the reference density $I_0$ as $\hat{I} = (f(x) - x) \sqrt{I_0(x)}$, then:

\begin{equation}
    \begin{aligned}
        \hat{I}_\sigma &= (f_\sigma(x) - x) \sqrt{I_0(x)}\\
        &= \frac{1}{\sigma} \left(f_1(x) - x + x - \sigma x \right) \sqrt{I_0(x)}\\
        &= \frac{1}{\sigma} \left( \hat{I}_1(x) - x(\sigma - 1)\sqrt{I_0(x)} \right)
    \end{aligned}
\end{equation}
\section{Proof of the Composition Property}
\label{app:proof-composition}

Consider two positive density functions $I_0$ and $I_1$ defined on measure spaces $X \subseteq \RR^d$ and $Y \subseteq \RR^d$, respectively. We assume that the total mass associated with $X$ is equal to the total mass associated with $Y$, such that:

\begin{equation}
    \int_{X}I_0(x)dx = \int_{Y}I_1(y)dy
\end{equation}
The optimal transport problem is to find a mass-preserving map $f_1 : X \rightarrow Y$ that maps one density to the other:

\begin{equation}
    \int_{A} I_0(x)dx = \int_{f_1(A)} I_1(y)dy \quad \forall A \subset X
    \label{eq:composition-I1-map}
\end{equation}
If we consider $f_1$ to be the gradient of a convex potential $\phi_1$, such that $f_1 = \nabla \phi_1$, then by Brenier's theorem, there exists a unique (up to a constant) transport map $f_1$, that will satisfy the equation above. Now let $I_g$ represent the composition of $I_1$ with an invertible function $g(y)$, such that $I_g(y) = D_g(y)I_1(g(y))$ where $D_g$ is the determinant of the Jacobian of $g$. Similarly to $f_1$, $g$ is also the gradient of a potential $\varphi_g$ (i.e. $g = \nabla \varphi_g$). In the same manner as before, we can solve for $f_g : X \rightarrow Y$:

\begin{equation}
    \begin{aligned}
        \int_{A} I_0(x)dx &= \int_{f_g(A)} I_g(y)dy\\
        &= \int_{f_g(A)} D_g(y)I_1(g(y))dy
    \end{aligned}
    \label{eq:composition-Imu-map}
\end{equation}
By substituting $u = g(y)$ into Eq. \eqref{eq:composition-Imu-map} via the change of variables theorem, and equating Eqs. \eqref{eq:composition-I1-map} and \eqref{eq:composition-Imu-map} we get:

\begin{equation}
    \int_{f_1(A)} I_1(y)dy = \int_{g(f_g(A))} I_1(u)du
\end{equation}
Since the integrals are equal, we have $f_g(x) = g^{-1}(f_1(x))$. If we define the transform of density $I$ with respect to the reference density $I_0$ as $\hat{I} = (f(x) - x) \sqrt{I_0(x)}$, then:

\begin{equation}
    \begin{aligned}
        \hat{I}_g &= (f_g(x) - x) \sqrt{I_0(x)}\\
        &= (g^{-1}(f_1(x)) - x) \sqrt{I_0(x)}\\
        &= \left( g^{-1} \left( \frac{\hat{I}_1(x)}{\sqrt{I_0(x)}} + x \right) - x \right) \sqrt{I_0(x)}
    \end{aligned}
\end{equation}
\section{Proof of the Linear Separability Property}
\label{app:proof-linear-separability}

Two non-empty subsets $\hat{\PP}$ and $\hat{\QQ}$ of a vector space $\Omega$ are linearly separable if, and only if, their convex hulls are disjoint, i.e:

\begin{equation}
    \sum_{i=1}^{N_p} \alpha_i \hat{p}_i \neq \sum_{j=1}^{N_q} \beta_j \hat{q}_j
    \label{eq:convex-hull-disjoint}
\end{equation}
for any subset $\{\hat{p}\}_{i=1}^{N_p} \subset \hat{\PP}$ and $\{\hat{q}\}_{j=1}^{N_q} \subset \hat{\QQ}$, where $\alpha_i, \beta_j > 0$ and $\sum_i \alpha_i = \sum_j \beta_j = 1$. For a proof, we refer the reader to \cite{park2017}.

In order to show that the LOT transforms $\hat{\PP}$ and $\hat{\QQ}$ (as defined in Eq. \eqref{eq:lot_forward}) of sets $\PP$ and $\QQ$ are linearly separable, we first assume that $\hat{\PP}$ and $\hat{\QQ}$ are not linearly separable (proof by contradiction). This implies that the convex hulls of the LOT transforms are not disjoint, and therefore, from Eq. \eqref{eq:convex-hull-disjoint} we can write:

\begin{equation}
    \label{eq:joint_sets_lot}
    \sum_i \alpha_i \hat{p}_i = \sum_j \beta_j \hat{q}_j
\end{equation}
Since the LOT transforms are defined as $\hat{p}_i = (f_i - x)\sqrt{I_0}$ and $\hat{q}_j = (g_j - x)\sqrt{I_0}$, the equation above can be rewritten in terms of the transport maps $f_i$ and $g_j$:

\begin{equation}
    \begin{aligned}
        \sum_i \alpha_i (f_i - x)\sqrt{I_0} &= \sum_j \beta_j (g_j - x)\sqrt{I_0}\\
        \sum_i \alpha_i f_i &= \sum_j \beta_j g_j.
    \end{aligned}
    \label{eq:separability-fi-gj}
\end{equation}
where, by Brenier's theorem, $f_i = \nabla \phi_i$ and $g_j=\nabla \varphi_j$, and $\phi_i, \varphi_j$ are convex potentials.

Using the Jacobian equation from the Monge formulation of the optimal transport problem (Eq. \eqref{eq:jacobian-equation}), and the fact that the LOT transform uniquely identifies an $f_0$ with some $p_0$, we can write $D_{f_0} p_0 \circ f_0 = I_0$. Moreover, using the generative model for image classes described in Section \ref{subsec:lot-linear-separability}, the elements $p_i$ of $\PP$ are generated via $p_i = D_{h_i} p_0 \circ h_i$. By combining these two results via the composition theorem for LOT transforms, one can show that $f_i = h_i^{-1} \circ f_0$. Similarly, one can also show that $g_j = h_j^{-1} \circ g_0$. These expressions for $f_i$ and $g_j$ can then be substituted into Eq. \eqref{eq:separability-fi-gj}:

\begin{equation}
    \begin{aligned}
        \sum_i \alpha_i h_i^{-1} \circ f_0 &= \sum_j \beta_j h_j^{-1} \circ g_0\\
        h_\alpha^{-1} \circ f_0 &= h_\beta^{-1} \circ g_0
    \end{aligned}
    \label{eq:separability-halpha}
\end{equation}
where $h_\alpha^{-1} = \sum_i \alpha_i h_i^{-1}$ and $h_\beta^{-1} = \sum_j \beta_j h_j^{-1}$ (permitted by conditions i and ii in Section \ref{subsec:lot-linear-separability}).

From Eq. \eqref{eq:jacobian-equation} we know that $D_{f_0} p_0 \circ f_0 = D_{g_0} q_0 \circ g_0 = I_0$. If we rearrange Eq. \eqref{eq:separability-halpha} and substitute it into this result, we get:

\begin{equation}
    (D_{h_\alpha} D_{h_\beta^{-1}} D_{g_0}) p_0 \circ (h_\alpha \circ h_\beta^{-1} \circ g_0) = D_{g_0} q_0 \circ g_0
\end{equation}
If we let $h_{\alpha\beta} = h_\alpha \circ h_\beta^{-1}$ (permitted by condition iii in Section \ref{subsec:lot-linear-separability}), then this becomes:

\begin{equation}
    D_{h_{\alpha\beta}} p_0 \circ h_{\alpha\beta} = q_0
\end{equation}
However, this contradicts the definition in Section \ref{subsec:lot-linear-separability}, which states that:

\begin{equation}
    D_h p_0 \circ h \neq q_0 \quad \forall h \in \HH
\end{equation}

Therefore, the LOT transforms $\hat{\PP}$ and $\hat{\QQ}$ of classes $\PP$ and $\QQ$ must be linearly separable.
\section{Derivative of the SPOT Objective Function in 2D}
\label{app:spot-objective}

Given that this work is primarily concerned with mapping images to one another, we extend the result in Eq. \eqref{eq:spot-spatial-transformation} to two dimensions:

\begin{equation}
    \begin{bmatrix}f(x,y)\\g(x,y)\end{bmatrix} = \begin{bmatrix}x\\y\end{bmatrix} -\nabla\phi(x,y)
    \label{eq:breniers-theorem-2d}
\end{equation}
where $f$ and $g$ are the transport maps in the $x$ and $y$ directions, respectively.

We can also rewrite Eq. (\ref{eq:potential}) in two dimensions:

\begin{equation}
    \phi(x,y) = \sum_p \sum_q c(p,q)\rho(x-p,y-q)
    \label{eq:potential-2d}
\end{equation}

By combining Eqs. (\ref{eq:breniers-theorem-2d}) and (\ref{eq:potential-2d}), we can express the transport maps $f$ and $g$ in terms of the potential $\phi$ :

\begin{eqnarray*}
    f(x,y) &=& x - \sum_p \sum_q c(p,q) \rho_x(x-p,y-q)\\
    g(x,y) &=& y - \sum_p \sum_q c(p,q) \rho_y(x-p,y-q)
\end{eqnarray*}
where subscripts denote derivatives with respect to that variable (e.g. $\rho_x = \frac{\partial\rho}{\partial x}$ and $\rho_{xx} = \frac{\partial^2\rho}{\partial x^2}$). Consequently, the first derivatives of the transport maps can be expressed as follows:

\begin{eqnarray*}
    f_c(x,y) &=& -\rho_x(x-p,y-q)\\
    g_c(x,y) &=& -\rho_y(x-p,y-q)\\
    f_x(x,y) &=& 1 - \sum_p \sum_q c(p,q) \rho_{xx}(x-p,y-q)\\
    f_y(x,y) &=& -\sum_p \sum_q c(p,q) \rho_{xy}(x-p,y-q)\\
    g_x(x,y) &=& -\sum_p \sum_q c(p,q) \rho_{yx}(x-p,y-q)\\
    g_y(x,y) &=& 1 - \sum_p \sum_q c(p,q) \rho_{yy}(x-p,y-q)
\end{eqnarray*}

In two-dimensions, the SPOT objective function in Eq. (\ref{eq:spot-objective}) becomes:

\begin{equation*}
    \psi(c) = \frac{1}{2} \sum_m \sum_n (D(m,n) I_1(f(m,n),g(m,n)) - I_0(m,n))^2
    \label{eq:spot-objective-2d}
\end{equation*}
where $I_0$ and $I_1$ are images that have been normalized, such that the total mass in each image is the same. The derivative of the objective function with respect to $c$ is given by:

\begin{multline}
    \psi_c(c) = \sum_m \sum_n r(m,n)\\
    \bigg( \frac{\partial}{\partial c}\big[D(m,n)\big] I_1(f(m,n),g(m,n))\\
    + D(m,n)\frac{\partial}{\partial c}\big[I_1(f(m,n),g(m,n))\big] \bigg)
    \label{eq:psi-derivative}
\end{multline}

with

\begin{equation*}
    r(x,y) = D(x,y) I_1(f(x,y),g(x,y)) - I_0(x,y)
\end{equation*}

and

\begin{equation}
    \begin{aligned}
        D(x,y) &= 
        \begin{vmatrix}
            f_x(x,y) & f_y(x,y)\\
            g_x(x,y) & g_y(x,y)
        \end{vmatrix}\\
        &= f_x(x,y)g_y(x,y) - f_y(x,y)g_x(x,y)
    \end{aligned}
    \label{eq:det-jacobian}
\end{equation}

Using Eq. (\ref{eq:det-jacobian}) and the product rule, the first partial derivative in Eq. (\ref{eq:psi-derivative}) can be written:

\begin{multline}
    \frac{\partial}{\partial c}D(x,y) = -g_y(x,y)\rho_{xx}(x-p,y-q)\\
    - f_x(x,y)\rho_{yy}(x-p,y-q) + g_x(x,y)\rho_{xy}(x-p,y-q)\\
    + f_y(x,y)\rho_{yx}(x-p,y-q)
    \label{eq:psi-deriv-part1}
\end{multline}

Similarly, using the chain rule, the second the partial derivative in Eq. (\ref{eq:psi-derivative}) is:

\begin{multline}
    \frac{\partial}{\partial c}I_1(f(x,y),g(x,y))\ =\\
    -{I_1}_x(f(x,y),g(x,y))\rho_x(x-p,y-q)\\
    - {I_1}_y(f(x,y),g(x,y))\rho_y(x-p,y-q)
    \label{eq:psi-deriv-part2}
\end{multline}

By substituting Eqs. (\ref{eq:psi-deriv-part1}) and (\ref{eq:psi-deriv-part2}) into Eq. (\ref{eq:psi-derivative}), and simplifying the result, $\psi_c$ can be written in the following way:

\begin{multline*}
    \psi_c(c) = \Lambda_1 * \tilde{\rho}_{xy}(p,q) + \Lambda_2 * \tilde{\rho}_{yx}(p,q) - \Lambda_3 * \tilde{\rho}_{xx}(p,q)\\
    - \Lambda_4 * \tilde{\rho}_{yy}(p,q) - \Lambda_5 * \tilde{\rho}_x(p,q) - \Lambda_6 * \tilde{\rho}_y(p,q) 
\end{multline*}
where $*$ denotes convolution, a tilde above an array denotes flipping the array horizontally and vertically (i.e. $\tilde{\rho}(x,y) = \rho(-x,-y)$), and:

\begin{eqnarray*}
    \Lambda_1(x,y) &=& r(x,y)g_x(x,y)I_1(f(x,y),g(x,y))\\
    \Lambda_2(x,y) &=& r(x,y)f_y(x,y)I_1(f(x,y),g(x,y))\\
    \Lambda_3(x,y) &=& r(x,y)g_y(x,y)I_1(f(x,y),g(x,y))\\
    \Lambda_4(x,y) &=& r(x,y)f_x(x,y)I_1(f(x,y),g(x,y))\\
    \Lambda_5(x,y) &=& r(x,y)D(x,y){I_1}_x(f(x,y),g(x,y))\\
    \Lambda_6(x,y) &=& r(x,y)D(x,y){I_1}_y(f(x,y),g(x,y))
\end{eqnarray*}

\bibliography{references}

\end{document}